\documentclass{article}
\usepackage{tag_preprint,times}
\iclrfinalcopy

\usepackage{amsmath,amssymb,amsfonts,bm}
\usepackage{algorithm,algorithmic}
\usepackage{array,booktabs,multirow,threeparttable}
\usepackage{siunitx}
\usepackage{bbding}   %
\usepackage{textcomp,wasysym,cancel,verbatim}
\usepackage{url}

\usepackage[dvipsnames]{xcolor}

\usepackage[export]{adjustbox}
\usepackage{threeparttable}
\usepackage{multirow}
\usepackage{enumitem}
\usepackage{titletoc}
\usepackage{newfloat}
\usepackage{listings}
\usepackage{colortbl}
\usepackage{arydshln}
\usepackage{amsfonts}
\usepackage{wrapfig}
\usepackage{needspace}
\usepackage{placeins}
\usepackage{amssymb}
\usepackage{pifont}
\usepackage{float}
\usepackage{bbm}
\usepackage{bm}
\usepackage{mathrsfs}  
\usepackage{tabularx}
\usepackage{graphicx}
\usepackage{fontawesome}

\definecolor{mygray}{gray}{.9}
\definecolor{mygreen}{RGB}{93,173,85}
\definecolor{logocolor}{RGB}{0,207,205}
\definecolor{mycitecolor}{HTML}{3498DC}
\definecolor{mylinkcolor}{HTML}{E74D3B}
\definecolor{myurlcolor}{HTML}{980000}
\definecolor{mywarning}{RGB}{233,144,61}
\definecolor{mydarkgreen}{HTML}{6a994e}
\definecolor{green(pigment)}{rgb}{0.0, 0.65, 0.31}

\usepackage{hyperref}
\hypersetup{colorlinks=true,linkcolor=mylinkcolor,citecolor=mycitecolor,urlcolor=myurlcolor}

\definecolor{DarkBlue}{RGB}{64,101,149}
\definecolor{azure}{rgb}{0.0, 0.5, 1.0}
\definecolor{gray}{rgb}{0.3, 0.3, 0.3}
\definecolor{DarkGreen}{RGB}{42,110,63}
\definecolor{DarkYellow}{RGB}{191,144,0}
\definecolor{DarkRed}{rgb}{0.6, 0, 0}

\newcolumntype{x}[1]{>{\centering\arraybackslash}p{#1pt}}
\newcolumntype{I}{!{\vrule width 1pt}}

\usepackage[most]{tcolorbox}
\tcbuselibrary{theorems}
\definecolor{lightgray}{gray}{.9}
\definecolor{deepgray}{gray}{.8}
\tcbset{highlight math/.append style={left=0mm,right=0mm,top=0mm,bottom=0mm, colframe=white}}

\makeatletter
\newcommand{\thickhline}{%
    \noalign {\ifnum 0=`}\fi \hrule height 1pt
    \futurelet \reserved@a \@xhline
}



\newcommand{\xmark}{\ding{55}}%
\newcommand{\pub}[1]{{\color{gray}{\fontsize{4.5}{6}\selectfont {[{#1}]}}}}

\usepackage{xspace}
\makeatletter
\DeclareRobustCommand\onedot{\futurelet\@let@token\@onedot}
\def\@onedot{\ifx\@let@token.\else.\null\fi\xspace}
 
\def\ie{\textit{i.e}\onedot}

\definecolor{darksalmon}{rgb}{0.91, 0.59, 0.48}
\definecolor{emerald}{rgb}{0.31, 0.78, 0.47}
\definecolor{green(pigment)}{rgb}{0.0, 0.65, 0.31}
\definecolor{amaranth}{rgb}{0.9, 0.17, 0.31}
\definecolor{iris}{rgb}{0.35, 0.31, 0.81}
\definecolor{uu}{rgb}{0.95, 0.51, 0.51}
\definecolor{spirodiscoball}{rgb}{0.06, 0.75, 0.99}
\usepackage{bbding}

\definecolor{myblue}{HTML}{4594c1}
\definecolor{oursblue}{HTML}{4682b4}
\DeclareRobustCommand{\TAG}{\textnormal{\bfseries\itshape TAG}}
\DeclareRobustCommand{\TAGI}{\textnormal{\bfseries\itshape TAG-I}}
\DeclareRobustCommand{\TAGV}{\textnormal{\bfseries\itshape TAG-V}}
\hypersetup{
  pdftitle={TAG: Calibrating SAM3 for Universal Image Segmentation without Dense Supervision},
  pdfauthor={Hao Zhu, Heyu Sun, Shuo Jin, Jiayu Xiao, Feng Dai},
  pdfsubject={Universal image segmentation},
  pdfkeywords={SAM3, universal segmentation, weak supervision, vision-language models}
}

\graphicspath{{figures/}}
\renewcommand{\arraystretch}{1.4}

\title{\textbf{\textit{TAG}}: Calibrating SAM3 for Universal Image Segmentation without Dense Supervision}

\author{Hao Zhu, Heyu Sun, Shuo Jin, Jiayu Xiao, Feng Dai \\}
\preprintemail{zhuhao22z@ict.ac.cn}
\date{}

\begin{document}
\begingroup
\renewcommand{\arraystretch}{1}
\maketitle
\vspace{1.3pt}
\endgroup
\lhead{Preprint. Under review}

\begin{abstract}
The recent Segment Anything Model 3 (SAM3) extends powerful promptable segmentation from spatial cues to concepts, raising the prospect of a single foundation model for universal image segmentation.
However, this promise often fails to translate into reliable performance on downstream segmentation tasks. 
In this work, we empirically identify concept recognition, rather than mask quality, as the primary bottleneck: SAM3 accurately delineates the objects it recognizes yet struggles to robustly determine which concepts are present.
Driven by this, we present \textit{\underline{T}ext \underline{A}nchor re-\underline{G}rounding (\TAG{})}, a simple yet effective framework that rectifies the concept recognition of SAM3 on the target domain without dense annotations.
Specifically, \TAG{} introduces an \textit{Anchor Residual Pathway} comprising image-conditioned, low-rank adapters to re-ground text anchors in the target distribution and modulate subsequent text-conditioned interactions.
Guided solely by image-level category labels, \TAG{} jointly optimizes binary cross-entropy with \textit{Soft Boundary Ranking} objectives to calibrate presence confidence and distinguish present concepts from confusable negatives.
Building upon this, we devise \textit{Confidence-Weighted Distillation} to couple rich semantic context of a multimodal large language model with fine-grained spatial priors of SAM3, 
thereby achieving effective calibration in the absence of manual annotations.
Extensive experiments demonstrate that \TAG{} unlocks SAM3's segmentation potential with only 3.1M trainable parameters, achieving impressive performance across diverse scenarios, and notably, even surpassing leading mask-supervised methods on several metrics.
\href{https://github.com/MiSsU-HH/TAG_SAM3}{Our code will be publicly available at GitHub \faGithub}.
\end{abstract}

\section{Introduction}

Image segmentation has long stood as a fundamental pillar of visual understanding.
Recent advances increasingly pursue a unified model capable of handling semantic, instance, and panoptic segmentation~\citep{jain2023oneformer,maskformer}. 
To this end, prevailing universal frameworks~\citep{mask2former,cavagnero2024pem} unify the three tasks under a shared mask-classification paradigm, and successive improvements in model architecture and learning objectives have steadily refined mask quality~\citep{li2023maskdino,li2024omg}.
Despite remarkable progress, carrying such a model to new target domains remains challenging. 
On the one hand, adapting these models to a target domain relies heavily on dense pixel-level annotations, which are notoriously labor-intensive and time-consuming to obtain.
On the other hand, such specialized models typically require full fine-tuning for each deployment scenario, incurring substantial computational overhead.

\begin{figure}[t]
  \centering
  \includegraphics[width=\textwidth]{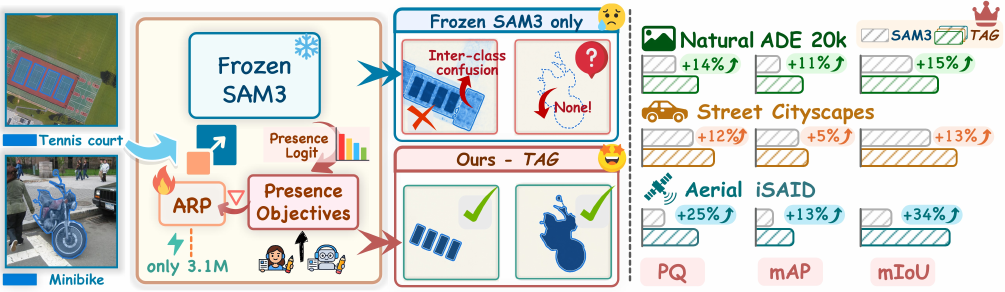}
  \par\vspace{-5pt}
  \caption{
  \textbf{\TAG{} calibrates SAM3 with image-level supervision.}
  \TAG{} learns an ARP with only 3.1M parameters using image-level presence objectives while keeping SAM3 fully frozen.
  The examples illustrate how \TAG{} corrects the inter-class confusion and under-activation of SAM3.
  Right panel reports the gains achieved with image-level ground-truth labels across different scenarios.
  }
  \par\vspace{-8pt}
  \label{fig:intro}
\end{figure}

The Segment Anything Models (SAM)~\citep{sam,sam2} offer a promising route forward by amortizing dense mask learning through large-scale pretraining, yet their reliance on geometric prompts leaves semantic recognition external to the model.
Recent SAM3~\citep{sam3} marks a pivotal leap by seamlessly fusing text into its visual pipeline, extending promptable paradigm from spatial cues to concepts.
In principle, a frozen SAM3 can operate as a universal image segmenter, yielding both semantic and instance masks in a single pass when guided merely by target class names.
In practice, however, SAM3 falls markedly short of the promise when directly applied on downstream segmentation benchmarks.
As shown in \textcolor{mylinkcolor}{Figure~\ref{fig:intro}}, SAM3 exhibits two prevalent failure modes: i) \textit{inter-class confusion}, where the model falsely activates on visually similar yet incorrect categories, and ii) \textit{under-activation}, 
where clearly visible instances remain undetected.
To address this, a natural way is to fine-tune SAM3 with pixel-level annotations so that it adapts to the target distribution~\citep{sam3}.
Though effective, this manner reintroduces the substantial computational burden that large-scale pretraining is intended to alleviate and requires prohibitive mask annotations, limiting its real-world deployment.
This raises a central question: \textit{How can SAM3 be calibrated at low computational and annotation cost?}

Answering the above question first requires pinpointing where SAM3 fails.
Toward this end, this paper conducts an in-depth analysis of the failure modes of SAM3 (detailed results are provided in \textcolor{mylinkcolor}{\S\ref{sec:failure}}).
Crucially, our empirical analysis reveals that these failures do not arise from scattered pixel-level noise, instead, they form large contiguous regions spanning entire objects, causing objects to be misclassified or missed entirely.
By contrast, SAM3 accurately delineates the objects it does recognize.
Even with SAM3's raw masks left unchanged, segmentation performance can be markedly improved simply by correcting its category assignments and activations, as shown in \textcolor{mylinkcolor}{Figure~\ref{fig:failure-summary}b}.
\textit{These findings converge on a clear diagnosis: the mask prior of SAM3 transfers effectively to the target domains, whereas concept-level recognition struggles to generalize.}

Motivated by this insight, we propose \textit{\textbf{Text Anchor re-Grounding (\TAG{})}}, a simple yet effective framework that rectifies the concept recognition of SAM3 on the target domain without mask annotations.
Specifically, \textbf{on the learnable modules}, \TAG{} introduces an \textit{Anchor Residual Pathway} (ARP) that leverages an image-conditioned, low-rank text feature adapter to dynamically re-ground class embeddings and propagates these corrections through subsequent text-conditioned interactions, with all pretrained SAM3 weights frozen.
Besides, \textbf{on the learning objectives}, \TAG{} jointly optimizes binary cross-entropy and \textit{Soft Boundary Ranking} solely by image-level category labels to calibrate concept presence and separate present categories from confusable negatives, correcting the semantic decisions that govern mask prediction.
\textbf{To transcend reliance on manual annotations}, we devise \textit{Confidence-Weighted Distillation} strategy that robustly harnesses category labels from multimodal large language model (MLLM), \ie, Qwen3.8~\citep{qwen38}, to guide the learning of the ARP on unlabeled images.
In this way, \TAG{} distills the off-the-shelf semantic knowledge of MLLM into lightweight adapters, bridging two foundations with complementary strengths.

Extensive experiments on multiple benchmarks spanning natural images, urban scenes, and aerial imagery demonstrate the effectiveness and efficiency of \TAG{}. 
Despite introducing negligible \textbf{3.1M} trainable parameters under coarse image-level presence labels, \TAGI{}/\TAGV{} (human/MLLM supervised) remarkably boost the frozen SAM3 by 17.2/14.0 PQ, 9.5/7.5 AP, and 20.8/16.8  mIoU across three benchmarks.
Both variants surpass all competing weakly supervised and open-vocabulary methods, \textit{and notably}, even outperform leading mask-supervised models on key metrics.

\textbf{Briefly put, our contributions are summarized as follows}:
\begin{itemize}
\vspace{-0.5em}
    \item[\ding{202}] \textbf{Bottleneck diagnosis:} Through empirical analysis and oracle interventions, we reveal that the performance bottleneck of SAM3 in downstream segmentation tasks is concept grounding rather than mask quality. 
    Correcting semantic bias while leaving the raw masks intact can unleash the powerful dense perception capability of SAM3.
    \item[\ding{203}] \textbf{Targeted solution:} Guided by the diagnosis, we propose \TAG{}, which calibrates the text-conditioned process of SAM3 through an \textit{Anchor Residual Pathway}, optimized with binary cross-entropy and \textit{Soft Boundary Ranking} objectives from image-level category labels.
    We further introduce Confidence-Weighted Distillation to robustly harness the global semantic recognition priors of MLLM, generalizing \TAG {} to the unlabeled regime.
    \item[\ding{204}] \textbf{Broad validation:} Extensive evaluations across natural, urban scene and aerial benchmarks demonstrate the superior efficacy and efficiency of \TAG{}. With only 3.1M learnable parameters, it achieves impressive performance in both weakly supervised and annotation-free settings, even surpassing leading fully supervised methods on key metrics.
\end{itemize}

\section{Preliminaries}
\label{sec:background}

\subsection{A Recap on SAM3}

We build on the image branch of \textit{SAM3} (omitting all video components), which segments visual concepts specified by noun phrases.
Given an input image \(I \in \mathbb{R}^{H \times W \times 3}\) and a benchmark-defined label set \(\mathcal{C}=\{c_1,\ldots,c_K\}\), we use the canonical class name \(c_k\) as the text prompt. 
The image is first mapped to unconditioned visual tokens $\mathbf{F}_v\in\mathbb{R}^{N_v\times d}$ by image encoder, while each class name is independently encoded into text embeddings $\mathbf{F}_t^{k}\in\mathbb{R}^{N_t\times d}$ by text encoder.

For each class $k$, the fusion encoder conditions the visual tokens on the text tokens through stacked cross-attention layers.
The fusion encoder is followed by a DETR-like~\citep{detr} decoder that transforms a set of object queries $\mathbf{O}\in\smash[t]{\mathbb{R}^{Q\times d}}$ into class-conditioned predictions.
A single learned \textit{presence token} is concatenated with the object queries $\smash[t]{\tilde{\mathbf{O}}}=[\mathbf{O;o}_{\text{pres}}]\in\smash[t]{\mathbb{R}^{(Q+1)\times d}}$ to predict whether a category appears in an image.
Each decoder block applies cross-attention from $\smash[t]{\tilde{\mathbf{O}}}$ to the text embeddings and the conditioned visual tokens in turn.
After decoder, the presence token produces the global presence score $\smash[t]{p_{\mathrm{pres}}^{k}} = \sigma(\mathrm{MLP}_{\mathrm{pres}}(\smash{\mathbf{o}_{\mathrm{pres}}}))$ for each text input.
Following the MaskFormer~\citep{maskformer}, \textit{SAM3} produces both semantic and instance segmentation masks for each class $k$ through a shared segmentation head. 
The conditioned features from the fusion encoder are used to generate the semantic segmentation mask, whereas instance segmentation additionally leverages the object queries $\smash[t]{\mathbf{O}}$ output by the decoder.

For dense prediction, we follow SegEarth-OV3~\citep{li2025segearthov3} and adopt the presence score $\smash[t]{p_{\mathrm{pres}}^{k}}$ to weight mask predictions during filtering and fusion to suppress non-salient categories.

\subsection{Where Zero-Shot SAM3 Fails}\label{sec:failure}

\begin{figure}[t]
  \centering
  \includegraphics[width=\textwidth]{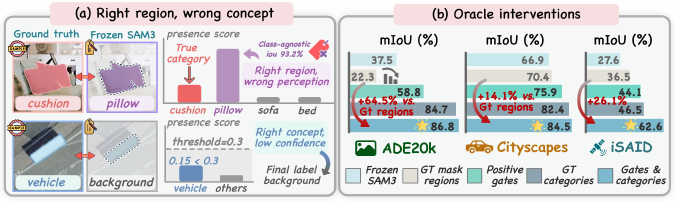}
  \par\vspace{-5pt}
  \caption{
  \textbf{Recognition errors in frozen SAM3.}
  \textbf{(a)} SAM3 exhibits strong foreground perception, yet its concept recognition often fails.
  \textbf{(b)} 
  Oracle interventions:
  \emph{Frozen SAM3} gives original predictions.
  \emph{GT mask regions} pairs GT mask regions with SAM3's original category scores.
  \emph{Positive gates} keeps raw SAM3 masks, setting only GT categories gate to $1$ for instance filtering.
  \emph{GT categories} pairs SAM3 masks with GT categories.
  \emph{Gates \& Categories} combines both semantic corrections.
  }
  \label{fig:failure-summary}
  \vspace{-8pt}
\end{figure}

Directly applied to downstream segmentation, SAM3 often confuses competing categories or misses visible objects.
In this work, we find that errors in SAM3 predominantly manifest as holistic, object-level failures, rather than the localized pixel-level noise commonly observed in traditional segmentation models.
Specifically, in \textcolor{mylinkcolor}{Figure~\ref{fig:failure-summary}a}, the mask predicted as \emph{pillow} closely follows the contour of \textit{cushion}, achieving \textbf{93.2\%} class-agnostic IoU with its ground-truth mask.
Despite this accurate delineation, the entire object is misclassified as a competing neighboring category.
A complementary failure occurs in the \textit{vehicle} example, where the confidence of the correct concept falls below the filtering threshold, causing the model to suppress the entire object as background.
Additional object-level statistics and diagnostic experiments are provided in \textcolor{mylinkcolor}{Appendix~\ref{sec:appendix-diagnosis}}.
These observations motivate the following hypothesis:
\textit{SAM3 exhibits robust spatial perception but fails to recognize concepts reliably, which prevents accurate object geometry from being translated into correct dense predictions and consequently leads to suboptimal performance on downstream benchmarks.}

\noindent\textbf{Oracle diagnosis of the recognition bottleneck.}
To investigate this hypothesis, we compare oracle interventions on region geometry, category assignment, and concept activation in the frozen model, as shown in \textcolor{mylinkcolor}{Figure~\ref{fig:failure-summary}b}.
Even perfect geometry (\emph{GT mask regions}) leaves substantial errors when regions are named by SAM3's  classification scores, and reduces accuracy on ADE20K.
With raw masks unchanged, \emph{GT categories} addresses inter-class confusion and improves ADE20K by \textbf{47.2} mIoU over zero-shot SAM3.
\emph{Positive gates} counters under-activation by enabling GT-present concepts and also improves accuracy.
Their combination remarkably outperforms \emph{GT regions} on every benchmark, demonstrating the larger benefit of correcting recognition over supplying perfect geometry.
These gains unlock the segmentation capacity already present in SAM3's spatial predictions but obscured by unreliable semantic recognition.

\noindent\textbf{Toward efficient adaptation of SAM3.}
Together, the object-level error pattern and the substantial gains from semantic correction support our hypothesis that SAM3 retains strong spatial priors for mask delineation, yet its category recognition fails to generalize reliably to downstream domains.
It suggests an efficient way for adapting SAM3. 
Rather than modifying its pretrained spatial representations, we can preserve all pretrained weights to reuse the spatial priors, and concentrate the learning effort on correcting \textit{category-level semantic recognition}. 
In this setting, higher-level semantic labels can provide effective supervision at relatively low annotation cost, while the pretrained model supplies the fine-grained spatial information required for dense prediction.

\section{Method}\label{sec:method}

\begin{figure}[t]
  \centering
  \includegraphics[width=\textwidth]{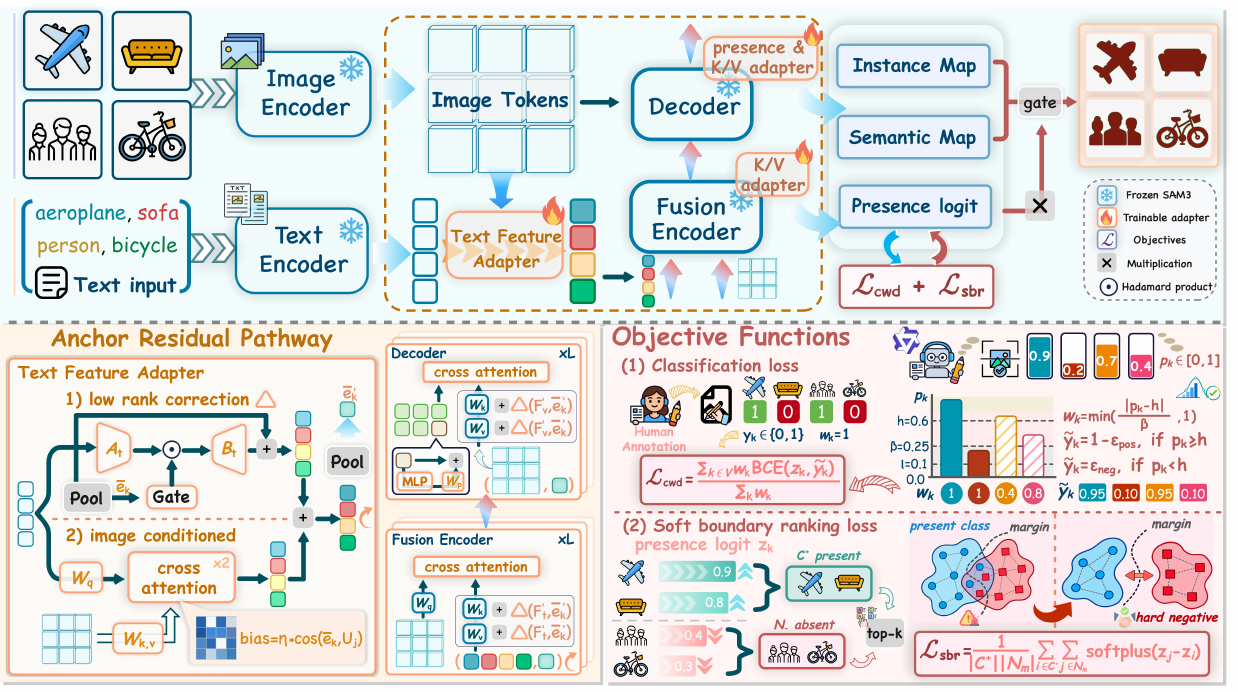}
  \par\vspace{-8pt}
  \caption{\textbf{Overview of our proposed \TAG{}.}
  Frozen SAM3 encodes the image and class prompts, while trainable adapters refine text features, cross-modal attention, and presence prediction using image-level supervision.
  Here, \includegraphics[height=1.5ex]{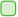} and \includegraphics[height=1.5ex]{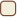} denote the object queries and presence token, respectively.
  }
  \label{fig:pipeline}
  \par\vspace{-8pt}
\end{figure}

We present \TAG{}, which re-grounds SAM3's semantic anchors in the target domain while keeping all pretrained weights frozen (\textcolor{mylinkcolor}{Figure~\ref{fig:pipeline}}).
We introduce the \textit{Anchor Residual Pathway} (\textcolor{mylinkcolor}{\S\ref{sec:arp}}) and its \textit{learning objectives} under image-level supervision (\textcolor{mylinkcolor}{\S\ref{sec:objective}}).
To generalize across unlabeled images, we develop \textit{Confidence-Weighted Distillation} (\textcolor{mylinkcolor}{\S\ref{sec:vlm}}) using presence labels from an off-the-shelf MLLM.

\subsection{Anchor Residual Pathway}\label{sec:arp}

\noindent\textbf{Motivation.}
Within SAM3, the concept recognition is mediated by the text-conditioned process. 
The text features are computed once from the canonical class names and condition both cross-modal interactions and presence prediction throughout the model.
However, since the text features remain static across all images, they merely inherit each category's prototypical semantics from pre-training rather than capture how it actually appears in the target domain.
Under domain shift, this mismatch leads to the failures observed in \textcolor{mylinkcolor}{Figure~\ref{fig:intro}}, \ie, the anchor lies too close to a neighboring category (inter-class confusion) or the anchor lies too far from its own category (under-activation).

\noindent\textbf{Re-grounding the anchors.}
Driven by this, we freeze all pretrained SAM3 weights and focus adaptation on text features and the pathways through which they condition the model.
We first learn a low-rank residual~\citep{lora} to adapt the text features to the target domain:
\begin{equation}\label{eq:template}
  \Delta_{\mathrm t}(\mathbf F_t^k\mid\bar{\mathbf e}_k)
  =\big((\mathbf F_t^k\mathbf A)\odot g(\bar{\mathbf e}_k)\big)\mathbf B,
  \qquad
  g(\bar{\mathbf e}_k)=1+\tanh\!\big(\mathrm{MLP}(\bar{\mathbf e}_k)\big),
\end{equation}
here $\bar{\mathbf e}_k\in\mathbb{R}^{1\times d}$ denotes the pooled text tokens,  $\mathbf A\in\mathbb{R}^{d\times r}$ and $\mathbf B\in\mathbb{R}^{r\times d}$ are learnable projections shared across classes, with the rank $r=16$.
The class-specific gate $g(\bar{\mathbf e}_k)\in\mathbb{R}^{1\times r}$ reweights the low-rank features by element-wise multiplication.

Since the low-rank residual is image-invariant, we add \textit{text-to-image cross-attention} to modulate text anchors with visual evidence.
Specifically, we first map frozen visual features $\mathbf F_v$ to visual keys and values $\mathbf U\in\mathbb{R}^{N_v\times d}$.
Their cosine similarities with the pooled text feature $\bar{\mathbf e}_k$, scaled by a learned factor $\eta$, form a class-specific bias $\mathbf a_k\in\mathbb{R}^{1\times N_v}$ added to attention logits.
We refine the original text queries $\mathbf F_t^k$ through two trainable cross-attention blocks and form the image-conditioned residual as:
\begin{equation}\label{eq:image-residual}
  \Delta_{\mathrm{img}}^k(I)
  =\operatorname{Proj}_{\mathrm{out}}\!\left(
    \operatorname{CrossAttention}(\mathbf F_t^k,\mathbf U;\mathbf a_k)
  \right),
\end{equation}
where $\operatorname{Proj}_{\mathrm{out}}$ maps the final output to an $N_t\times d$ residual.
Finally, we combine the domain correction and image-conditioned residual to jointly update the original text features:
\begin{equation}\label{eq:text-residual}
  \mathbf F_t^{\prime k}(I)
  =\mathbf F_t^k+\Delta_{\mathrm t}(\mathbf F_t^k\mid\bar{\mathbf e}_k)
  +\Delta_{\mathrm{img}}^k(I).
\end{equation}

\noindent\textbf{Propagating semantic corrections.}
Within the fusion encoder, the corrected anchors are shared across all blocks, whereas the visual representations conditioned on them evolve progressively with network depth. 
To adapt this, we further introduce anchor residuals into the textual keys and values of the fusion encoder blocks.
The mean corrected text feature $\smash[b]{\bar{\mathbf e}'_k}\in\mathbb{R}^{1\times d}$ conditions residual updates to the text keys and values $\mathbf K,\mathbf V\in\mathbb{R}^{N_t\times d}$ before image-to-text cross-attention block:
\begin{equation}\label{eq:fusion-kv}
  \mathbf K'=\mathbf K+\Delta_K(\mathbf F_t^{\prime k}(I)\mid\bar{\mathbf e}'_k),
  \qquad
  \mathbf V'=\mathbf V+\Delta_V(\mathbf F_t^{\prime k}(I)\mid\bar{\mathbf e}'_k),
\end{equation}
here $\Delta_K$ and $\Delta_V$ follow \textcolor{mylinkcolor}{Eq.~\ref{eq:template}}.
The projection and gating parameters are shared across classes.
In addition, we leverage the mean corrected text feature to provide the presence token with a class-specific initialization, facilitating its integration of category semantics:
\begin{equation}\label{eq:presence-token}
  \mathbf o_{\mathrm{pres}}^{\prime k}
  =\mathbf o_{\mathrm{pres}}+\mathrm{MLP}_{\mathrm p}(\bar{\mathbf e}'_k)\mathbf W_{\mathrm p},
\end{equation}
where $\mathbf W_{\mathrm p}\in\mathbb{R}^{G\times d}$ contains $G$ learnable vectors shared across classes, and $\mathrm{MLP}_{\mathrm p}$ predicts their combination weights from $\bar{\mathbf e}'_k$.
To guide object queries toward visual evidence of the intended category, each decoder layer similarly refines the image keys and values projected from the fusion output $\mathbf F_v^k\in\mathbb{R}^{N_v\times d}$.
These residuals follow \textcolor{mylinkcolor}{Eq.~\ref{eq:template}}, using $\mathbf F_v^k$ as input and $\bar{\mathbf e}'_k$ for gating.

\subsection{Learning from Image-Level Presence}\label{sec:objective}\label{sec:opt}\label{sec:inference}

\noindent\textbf{Motivation.} 
Pretrained spatial priors of SAM3 supply the cues for mask delineation, thus, the supervision required for learning can be \textbf{\textit{coarser}} than the desired output.
Inspired by this, we optimize ARP by class labels to recalibrate category recognition, avoiding labor-intensive dense annotations.

\noindent\textbf{Per-class presence learning.}
For each image $I$ and input class $k$, the decoded presence token produces a classification logit $z_k=\mathrm{MLP}_{\mathrm{pres}}(\smash{\mathbf{o}_{\mathrm{pres}}})$, supervised by the binary presence label $y_k\in\{0,1\}$.
We train ARP with standard binary cross-entropy (BCE) objective on presence logits:
\begin{equation}\label{eq:presence-bce}
  \mathcal L_{\mathrm{pres}}
  =-\frac{1}{|\mathcal V|}\sum_{k\in\mathcal V}
  \left[y_k\log\sigma(z_k)
  +(1-y_k)\log\!\bigl(1-\sigma(z_k)\bigr)\right],
\end{equation}
where $\mathcal V$ denotes the supervised class indices in the current benchmark.

\noindent\textbf{Limitation.} BCE calibrates each category independently, but does not explicitly penalize pairwise ranking errors between positive classes and the most confident negatives.
This distinction matters because presence scores modulate class-specific masks that subsequently compete during inference.

\noindent\textbf{Soft Boundary Ranking.}
To explicitly separate present concepts from confusable absent alternatives, we introduce Soft Boundary Ranking (SBR), a smooth pairwise ranking objective.
Let $\mathcal C^+=\{k\in\mathcal V:y_k=1\}$ and $\mathcal C^-=\{k\in\mathcal V:y_k=0\}$ denote the present and absent classes, respectively.
We select $m$ absent classes with the highest logits as hard negatives $\mathcal N_m$ and minimize 
\begin{equation}\label{eq:sbr}
  \mathcal L_{\mathrm{sbr}}
  =\frac{1}{|\mathcal C^+|\,|\mathcal N_m|}
  \sum_{i\in\mathcal C^+}\sum_{j\in\mathcal N_m}
  \operatorname{softplus}(z_j-z_i),
\end{equation}
where $\operatorname{softplus}(u)=\log(1+e^u)$ and $m=3$ by default.
Together with BCE objective, it calibrates both absolute presence confidence and relative semantic preference, refining the presence decisions that control pretrained mask filtering.
Finally, the overall objective is defined as:
\begin{equation}\label{eq:loss}
  \mathcal L_{\mathrm{total}}
  =\underbrace{\lambda_{\mathrm{pres}}\mathcal L_{\mathrm{pres}}}_{\text{Presence calibration}}
  +\underbrace{\lambda_{\mathrm{rank}}\mathcal L_{\mathrm{sbr}}}_{\text{Concept ranking}}.
\end{equation}

\subsection{Semantic Distillation from a Vision-Language Model}\label{sec:vlm}

Learning ARP requires knowing which categories are present, rather than annotating their spatial extent.
This image-level supervision aligns with the global semantic understanding of multimodal large language models (MLLMs), complementing SAM3’s pretrained spatial priors. 
Motivated by this, we develop an \textit{unlabeled} variant by replacing human-annotated presence labels with MLLM-generated pseudo labels, enabling SAM3 to be recalibrated on unlabeled target images.

\noindent\textbf{Generating presence pseudo-labels.}
We query Qwen3.8-27B~\citep{qwen38} offline to identify the categories present in each training image. An example prompt is provided below:

\begingroup
\definecolor{presenceprompt}{RGB}{255,155,133}
\begin{tcolorbox}[
  enhanced,
  colback=presenceprompt!15!white, colframe=presenceprompt,
  colbacktitle=presenceprompt!90!white, coltitle=black,
  title={Presence labeling prompt (summary)},
  fonttitle=\small\bfseries, fontupper=\small,
  boxrule=0.45pt, arc=1.2mm,
  left=6pt, right=6pt, top=4pt, bottom=4pt,
  before skip=6pt, after skip=6pt
]
This image depicts \texttt{\{scene description\}}. Examine the entire image and identify which categories in \texttt{\{category list\}} are present, considering both foreground objects and background regions.\par\smallskip
Return only a JSON object with the listed category names as keys and \texttt{"yes"} or \texttt{"no"} as values.
\end{tcolorbox}
\endgroup

The resulting pseudo-labels for each image are continuous presence scores $p_k\in[0,1]$ for each class $k$, derived from the answer-token probabilities.
However, optimizing the naive BCE loss on binarized MLLM outputs discards critical confidence cues by penalizing ambiguous decisions on par with confident ones. In this setting, missed positives suppress valid concepts, while hallucinated positives can reinforce spurious presence predictions. 

\noindent\textbf{Confidence-Weighted Distillation.}
To mitigate this issue, we introduce Confidence-Weighted Distillation (CWD) strategy, which attenuates the influence of noisy supervision through confidence-aware weighting. 
CWD downweights uncertain pseudo-labels while preserving high-confidence cues for both concept presence and absence.
Given thresholds \(0\le \ell<h\le1\), we assign a positive binary target when \(p_k\ge h\) and a negative target otherwise, followed by separate smoothing:
\begin{equation}\label{eq:smoothing}
  \hat y_k=\mathbbm{1}[p_k\ge h],\qquad
  \tilde y_k=(1-\epsilon_{\mathrm{pos}})\hat y_k
  +\epsilon_{\mathrm{neg}}(1-\hat y_k),
\end{equation}
here $\mathbbm{1}[\cdot]$ denotes the indicator function, $\hat y_k$ and $\tilde y_k$ are the binarized target and its smoothed counterpart. 
The parameters $\epsilon_{\mathrm{pos}},\epsilon_{\mathrm{neg}}\in[0,1/2)$ independently control the smoothing strength for positive and negative targets, respectively.
We reduce the weights of pseudo-labels with presence scores close to the decision threshold $h$ in the BCE objective function, formulated as:
\begin{equation}\label{eq:cwd}
  w_k=\min\!\left\{1,\frac{|p_k-h|}{\beta}\right\},\qquad
  \mathcal L_{\mathrm{cwd}}
  =\frac{\sum_{k\in\mathcal V}w_k\,\mathrm{BCE}(z_k,\tilde y_k)}
  {\sum_{k\in\mathcal V}w_k},
\end{equation}
where $w_k\in[0,1]$ denotes the confidence weight for class $k$, and $\beta=\max\{(h-\ell)/2,10^{-6}\}$ is the distance scale.
We replace $\mathcal L_{\mathrm{pres}}$ in \textcolor{mylinkcolor}{Eq.~\ref{eq:loss}} with $\mathcal L_{\mathrm{cwd}}$.
Meanwhile, SBR uses only the confident classes $\mathcal V_{\mathrm{conf}}=\{k\in\mathcal V:p_k\le\ell\ \text{or}\ p_k\ge h\}$, forming both positive set and hard-negative candidates from this set according to $\hat y_k$.
Intuitively, \TAG{} can be interpreted as a form of semantic distillation, transferring the semantic understanding of MLLM to mitigate SAM3's recognition bottleneck.

\newcommand{\tagI}{\textcolor{oursblue}{\TAGI} (Ours)}
\newcommand{\tagV}{\textcolor{oursblue}{\TAGV} (Ours)}
\definecolor{oursrow}{HTML}{D7F6FF}
\definecolor{oursrowsoft}{HTML}{EAFBFF} %

\providecommand{\redup}[1]{{\scriptsize\color{RedOrange}$\uparrow$\hspace{-1pt}#1}}
\providecommand{\bluedown}[1]{{\scriptsize\color{BlueGreen}$\downarrow$\hspace{-1pt}#1}}
\newcommand{\gnum}[1]{\textcolor{gray}{#1}}
\newcommand{\tbd}{\textcolor{red}{\scriptsize TBD}}
\newcommand{\best}[1]{\textbf{#1}}
\newcommand{\snd}[1]{\underline{#1}}

\newcommand{\supP}{$\mathcal{M}$}
\newcommand{\supI}{$\mathcal{I}$}
\newcommand{\supPt}{$\mathcal{P}$}   %
\newcommand{\supV}{$\mathcal{V}$}
\newcommand{\supN}{\xmark}
\newcommand{\supMs}{$\widetilde{\mathcal{M}}$}

\newcommand{\repro}{$^{\dag}$}
\newcommand{\reeval}{$^{\ddag}$}
\section{Experiments}\label{sec:experiments}

\begin{table}[!t]\small
\centering
\setlength{\abovecaptionskip}{0pt}
\caption{\textbf{Universal segmentation comparison} on three validation benchmarks.
Sup.: \supP{} target masks, \smash[t]{\supMs{}} source masks, \supI{} image labels, \supPt{} points, \supV{} MLLM labels; \#Par.: trainable parameters.
$\uparrow$: gains over frozen SAM3. See \textcolor{mylinkcolor}{Appendix~\ref{sec:appendix-counterparts}} for more details.}
\vspace{2pt}
\renewcommand{\redup}[1]{{\tiny\color{RedOrange}$\uparrow$#1}}
\resizebox{\textwidth}{!}{%
    \setlength\tabcolsep{1.2pt}
    \renewcommand\arraystretch{1.15}
    \begin{tabular}{rccc!{\vrule width 0.8pt}ccc!{\vrule width 0.4pt}ccc!{\vrule width 0.4pt}ccc}
    \noalign{\hrule height 1.2pt}
    \rowcolor{gray!20} & & & & \multicolumn{3}{c!{\vrule width 0.4pt}}{\textbf{ADE20K}} & \multicolumn{3}{c!{\vrule width 0.4pt}}{\textbf{Cityscapes}} & \multicolumn{3}{c}{\textbf{iSAID}} \\
    \cline{5-13}
    \rowcolor{gray!20}
    \multicolumn{1}{c}{\multirow{-2}{*}{\textbf{Method}}} & \multirow{-2}{*}{\textbf{Sup.}} & \multirow{-2}{*}{\textbf{Backbone}} & \multirow{-2}{*}{\textbf{\#Par.}}
    & PQ & AP & mIoU & PQ & AP & mIoU & PQ & AP & mIoU \\
    \hline\hline
    \rowcolor{white}
    \multicolumn{13}{l}{\textcolor{gray!60}{\textit{Fully supervised universal segmentation}}} \\
    \rowcolor{white}
    Mask2Former~\pub{CVPR22} & \supP & Swin-L & 216M
    & 48.1 & 34.9 & 56.1 & 66.6 & 43.7 & 83.3 & 58.1 & 39.5 & 68.5 \\
    \rowcolor{gray!10}
    MaskDINO~\pub{CVPR23} & \supP & R50 & 52M
    & -- & -- & 48.7 & -- & -- & 80.0 & 57.4 & 36.8 & 64.6 \\
    \rowcolor{white}
    OneFormer~\pub{CVPR23} & \supP & DiNAT-L & 223M
    & 50.5 & 36.0 & 58.3 & 67.6 & 45.6 & 83.1 & 51.3 & 31.7 & 62.9 \\
    \rowcolor{gray!10}
    PEM~\pub{CVPR24} & \supP & R50 & 35.6M
    & 38.5 & -- & 45.5 & 61.1 & -- & 79.9 & 49.1 & 30.7 & 59.7 \\
    \rowcolor{white}
    EoMT~\pub{CVPR25} & \supP & DINOv2-L & 316M
    & 50.6 & -- & 58.4 & -- & -- & 84.2 & 56.7 & 36.3 & 69.7 \\
    \rowcolor{gray!10}
    ViT-P~\pub{CVPR26} & \supP & InternImage-H & 1.4B
    & 54.5 & 40.6 & 61.2 & 70.8 & 50.6 & 85.4 & -- & -- & -- \\
    \hline\hline
    \rowcolor{white}
    \multicolumn{13}{l}{\textcolor{gray!60}{\textit{Open-vocabulary universal segmentation}}} \\
    \rowcolor{white}
    X-Decoder~\pub{CVPR23} & \supMs & DaViT-L & 340.5M
    & 21.8 & 13.1 & 29.6 & 38.1 & 24.9 & 52.0 & 28.1 & 17.6 & 37.1 \\
    \rowcolor{gray!10}
    HIPIE~\pub{NeurIPS23} & \supMs & ViT-H & 202.3M
    & 22.9 & 19.0 & 29.0 & -- & -- & -- & 28.6 & 13.2 & 9.5 \\
    \rowcolor{white}
    APE~\pub{CVPR24} & \supMs & ViT-L & 380.3M
    & 27.2 & 24.4 & 30.0 & 33.3 & 30.3 & 44.2 & 32.9 & 23.5 & 37.4 \\
    \rowcolor{gray!10}
    SAM-CP~\pub{ICLR25} & \supMs & ConvNeXt-L CLIP & --
    & 27.2 & 17.0 & 31.8 & 41.0 & 29.3 & 47.9 & -- & -- & -- \\
    \rowcolor{white}
    OpenWorldSAM~\pub{NeurIPS25} & \supMs & SAM2{\,+\,}VLM & 4.5M
    & 35.2 & 16.9 & 60.4 & 36.1 & 12.0 & 57.4 & 21.6 & 10.5 & 22.3 \\
    \rowcolor{gray!10}
    FrozenSeg~\pub{IJCV26} & \supMs & ConvNeXt-L{\,+\,}SAM & 26.5M
    & 25.9 & 16.4 & 34.4 & 45.8 & 28.4 & 56.8 & 33.5 & 19.2 & 32.5 \\
    \hline\hline
    \rowcolor{white}
    \multicolumn{13}{l}{\textcolor{gray!60}{\textit{Weakly supervised universal segmentation}}} \\
    \rowcolor{white}
    Point2Mask~\pub{ICCV23} & \supPt & R50 & 56.5M
    & 18.6 & 11.8 & 26.3 & 21.2 & 3.5 & 39.9 & 5.2 & 1.0 & 31.9 \\
    \hdashline
    \rowcolor{gray!10}
    SAM3~\pub{ICLR2026} & - & SAM3 & \textit{freeze}
    & 32.7 & 25.0 & 37.5 & 50.3 & 36.5 & 66.9 & 19.2 & 12.2 & 27.6 \\
    \rowcolor{oursrowsoft}
    \tagV{} & \supV{} & SAM3 & 3.1M
    & \textbf{40.8}\redup{8.1} & \textbf{31.5}\redup{6.5} & \textbf{47.0}\redup{9.5} & \textbf{61.6}\redup{11.3} & \textbf{41.0}\redup{4.5} & \textbf{78.1}\redup{11.2} & \textbf{41.9}\redup{22.7} & \textbf{23.6}\redup{11.4} & \textbf{57.4}\redup{29.8} \\
    \rowcolor{oursrow}
    \tagI{} & \supI & SAM3 & 3.1M
    & \best{46.9}\redup{14.2} & \best{35.7}\redup{10.7} & \best{52.7}\redup{15.2} & \best{62.7}\redup{12.4} & \best{41.0}\redup{4.5} & \best{80.3}\redup{13.4} & \best{44.2}\redup{25.0} & \best{25.5}\redup{13.3} & \best{61.4}\redup{33.8} \\
    \noalign{\hrule height 1.2pt}
    \end{tabular}
}
\label{tab:universality}
\end{table}

\noindent\textbf{Benchmarks.}\label{sec:benchmarks}
We conduct our main experiments on three benchmarks spanning natural images \textbf{ADE20K}~\citep{ade20k}, urban street scenes \textbf{Cityscapes}~\citep{cityscapes}, and aerial imagery \textbf{iSAID}~\citep{isaid}.
Compared with Pascal VOC~\citep{voc} and COCO~\citep{coco_obj}, two widely used segmentation benchmarks, we prioritize the more challenging datasets described above to enable a rigorous evaluation of cross-domain adaptation. 
More comparative results on Pascal VOC and COCO are provided in \textcolor{mylinkcolor}{Appendix~\ref{sec:appendix-weak}}.
Following standard universal image segmentation evaluation~\citep{mask2former}, we report panoptic quality (PQ)~\citep{kirillov2019panoptic}, mask average precision (AP), and mean intersection-over-union (mIoU) for panoptic, instance, and semantic segmentation tasks, respectively.

\noindent\textbf{Implementation details.}\label{sec:implementation}
Our work builds upon the pretrained SAM3~\citep{sam3}, which employs Perception Encoder ~\citep{bolya2025perception} for image and text encoding.
As for text input, we directly use the canonical class names defined by each benchmark as concept prompts.
We adopt the open-source Qwen3.8 27B~\citep{qwen38} as the default MLLM for generating labels in unlabeled setting.
Our implementation is based on $\mathtt{Pytorch}$, and all experiments are conducted on NVIDIA RTX 4090 GPUs.
More implementation details are provided in \textcolor{mylinkcolor}{Appendix~\ref{sec:appendix-method}}.

\subsection{Comparative results}

\noindent\textbf{Counterparts.}\label{sec:counterparts}
Achieving universal image segmentation under weak supervision represents a formidable yet nascent frontier.
While pioneering efforts~\citep{shen2021toward,li2023point2mask} have made commendable strides, their efficacy remains limited on challenging scenarios.
In light of this, we expand the comparison to leading fully supervised and open-vocabulary methods, positioning \TAG{} within the broader landscape of segmentation research (see \textcolor{mylinkcolor}{Appendix~\ref{sec:appendix-counterparts}} for more details).

\noindent\textbf{Strong priors make weak supervision effective.}\label{sec:results-weak}
\textcolor{mylinkcolor}{Table~\ref{tab:universality}} compares \TAG{} with weakly supervised methods on challenging cross-domain datasets.
Despite utilizing point annotations, these methods still remain suboptimal, whereas \TAG{} achieves impressive results relying solely on image-level presence.
\TAG{} inherits the spatial priors of SAM3 and directs weak supervision solely toward category recognition, providing a viable way for efficient and effective universal image segmentation.

\noindent\textbf{Optimal trade-off for application.}\label{sec:results-ov}
While open-vocabulary models can generalize to novel categories, they fail under severe domain shifts (\textcolor{mylinkcolor}{Appendix~\ref{sec:appendix-ov-transfer}}).
These models rely on mask-supervised adaptation within the related domain, which substantially limits the deployment advantages offered by vocabulary flexibility.
With only MLLM-generated labels, \TAGV{} surpasses all open-vocabulary counterparts across nine metrics, except the mIoU of OpenWorldSAM on ADE20K.
For tasks with predefined vocabularies, these results favor \TAG{} as a more robust and practical alternative.

\noindent\textbf{Lightweight adaptation narrows the performance gap.}
\label{sec:results-full}
With only 3.1M trainable parameters, our method substantially narrows the performance gap between SAM3 and leading fully supervised methods.
Notably, \TAGI{} even surpasses several counterparts, achieving 35.7 AP on ADE20K ahead of Mask2Former (\textit{vs.}34.9), while attaining 61.4 mIoU on iSAID over PEM (\textit{vs.}59.7).
These results demonstrate that our method is highly competitive and establishes a strong baseline. 
Future advances hold the potential to match or even exceed fully supervised SOTA under weak supervision.

\begin{figure}[t]
  \centering
  \includegraphics[width=\textwidth]{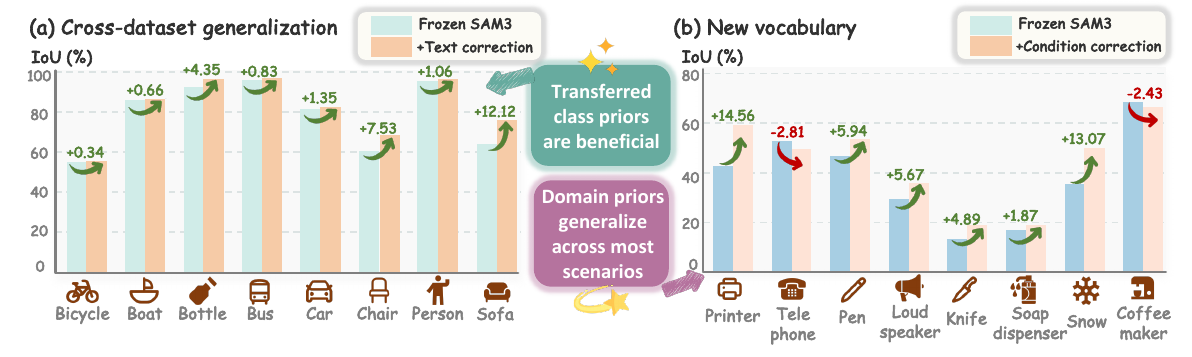}
  \par\vspace{-10pt}
  \caption{
  \textbf{Transferability of \TAG{}.}
  (a) Text residuals trained on ADE20K and generalized to shared VOC21 classes.
  (b) Class-agnostic adapters trained on ADE-150 and transferred to new classes.
  }
  \label{fig:general}
  \par\vspace{-10pt}
\end{figure}

\begin{table}[!t]
\centering
\small
\renewcommand{\arraystretch}{1.1}
\setlength{\tabcolsep}{1pt}
\begin{minipage}[b]{0.485\linewidth}
    \caption{\textbf{Learnable module ablations.}
    PQ (\%); w/o: without.\strut}
    \label{tab:ablation-modules}
\end{minipage}\hfill
\begin{minipage}[b]{0.485\linewidth}
    \caption{\textbf{Objective function ablations.}
    PQ (\%) for \TAGI{} and \TAGV{}.\strut}
    \label{tab:ablation-losses}
\end{minipage}\par
\nointerlineskip
\vspace{2pt}
\noindent
\begin{minipage}[t]{0.485\linewidth}
  \centering
  \fontsize{8}{10}\selectfont
  \begin{tabular}{p{\dimexpr.42\linewidth-12\tabcolsep-1.2pt\relax}!{\vrule width 0.8pt}*{3}{>{\centering\arraybackslash}p{.115\linewidth}}!{\vrule width 0.4pt}>{\centering\arraybackslash}p{.115\linewidth}>{\centering\arraybackslash}p{.12\linewidth}}
    \noalign{\hrule height 1.2pt}
    \rowcolor{lightgray}
    Setting & ADE & City & iSAID & \textbf{Avg.} & $\Delta$ \\
    \hline\hline
    \rowcolor{white}
    w/o feature adapter & 43.88 & 59.81 & 42.11 & 48.60 & \textcolor{DarkYellow}{$-2.66$} \\
    \rowcolor{white}
    w/o fusion K/V & 44.52 & 60.47 & 42.34 & 49.11 & \textcolor{DarkYellow}{$-2.15$} \\
    \rowcolor{white}
    w/o decoder K/V & 45.18 & 61.38 & 42.62 & 49.73 & \textcolor{DarkYellow}{$-1.53$} \\
    \rowcolor{white}
    w/o presence token & 45.99 & 61.36 & 43.74 & 50.36 & \textcolor{DarkYellow}{$-0.89$} \\
    \hdashline
    \rowcolor{oursrow}
    \textcolor{oursblue}{\textit{\textbf{All adapters}}} & \textbf{46.86} & \textbf{62.67} & \textbf{44.24} & \textbf{51.26} & -- \\
    \noalign{\hrule height 1.2pt}
  \end{tabular}
\end{minipage}\hfill
\begin{minipage}[t]{0.485\linewidth}
  \centering
  \fontsize{8}{10}\selectfont
  \begin{tabular}{p{\dimexpr.42\linewidth-12\tabcolsep-1.2pt\relax}!{\vrule width 0.8pt}*{3}{>{\centering\arraybackslash}p{.115\linewidth}}!{\vrule width 0.4pt}>{\centering\arraybackslash}p{.115\linewidth}>{\centering\arraybackslash}p{.12\linewidth}}
    \noalign{\hrule height 1.2pt}
    \rowcolor{lightgray}
    Setting & ADE & City & iSAID & \textbf{Avg.} & $\Delta$ \\
    \hline\hline
    \rowcolor{white}
    \hspace{0.6em}w/o SBR & 44.82 & 60.25 & 41.85 & 48.97 & \textcolor{DarkYellow}{$-2.28$} \\
    \rowcolor{white}
    \hspace{0.6em}w/o BCE & 43.48 & 57.77 & 41.66 & 47.64 & \textcolor{DarkYellow}{$-3.62$} \\
    \rowcolor{oursrow}
    \textcolor{oursblue}{\TAGI{}} & \textbf{46.86} & \textbf{62.67} & \textbf{44.24} & \textbf{51.26} & -- \\
    \hdashline
    \rowcolor{white}
    \hspace{0.6em}naive BCE & 37.54 & 57.65 & 38.59 & 44.59 & \textcolor{DarkYellow}{$-3.52$} \\
    \rowcolor{oursrowsoft}
    \textcolor{oursblue}{\TAGV{}} & \textbf{40.80} & \textbf{61.59} & \textbf{41.95} & \textbf{48.11} & -- \\
    \noalign{\hrule height 1.2pt}
  \end{tabular}
\end{minipage}
\vspace{-8pt}
\end{table}

\subsection{Diagnostic Experiments}\label{sec:ablation}

\noindent\textbf{Key Trainable Component Analysis.}
\textcolor{mylinkcolor}{Table~\ref{tab:ablation-modules}} shows that all four components contribute, as removing any one lowers PQ across the three datasets.
The largest degradation without the feature adapter highlights the central role of semantic anchor correction.
K/V and presence-token adapters further improve how these anchors guide cross-modal interaction and concept activation.
Their combined effectiveness supports ARP as a coordinated pathway for semantic re-grounding.

\noindent\textbf{Objective Functions.}
\textcolor{mylinkcolor}{Table~\ref{tab:ablation-losses}} supports the complementary roles of BCE and SBR.
SBR alone still improves over frozen SAM3, suggesting that relative concept ranking can partially alleviate its recognition bias.
Combining both objectives yields the best results, coupling absolute presence confidence with relative concept ordering.
For \TAGV{}, CWD consistently outperforms naive BCE, highlighting the importance of accounting for label uncertainty when transferring MLLM semantics.

\noindent\textbf{Generalization Analysis.} To further investigate the generalization of \TAG{}'s learned semantic priors, we conduct two transfer experiments.
Specifically, we \textit{i)} directly transfer the learned class-specific corrections to related domains, and \textit{ii)} reuse the class-agnostic parameters with an expanded vocabulary.
The quantitative results are reported in \textcolor{mylinkcolor}{Figure~\ref{fig:general}}.
\noindent\textbf{As for cross-dataset generalization,}
ADE20K-trained text residuals improve all shared VOC21 classes, with gains ranging from modest to pronounced.
\textbf{As for vocabulary expansion,} 
reusing the image-conditioned, fusion, and decoder adapters trained on ADE150 yields substantial gains for most of the eight newly added ADE847 categories.
These observations reveal that \textit{i)} the category corrections can capture recognition biases shared across related domains, and \textit{ii)} class-agnostic adapters learn visual priors shared across categories within the domain.
However, given that these gains are inconsistent and  occasionally negative, we argue that efficient adaptation to the target domain remains the most reliable choice.

\begin{figure}[t]
  \centering
  \includegraphics[width=\textwidth]{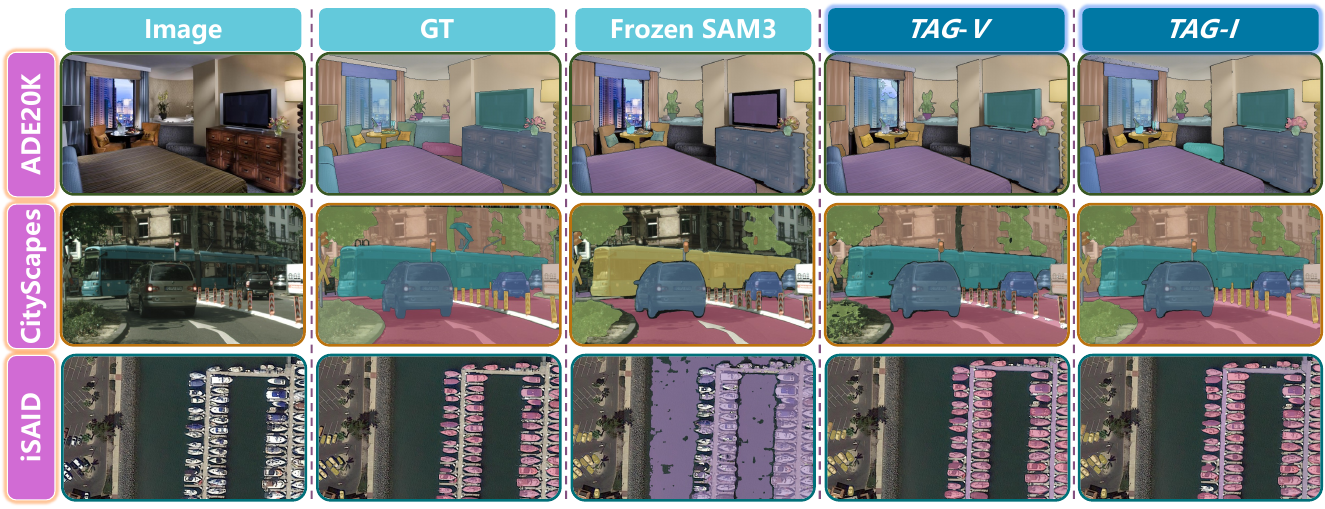}
  \par\vspace{-5pt}
  \caption{
    \textbf{Qualitative comparison} results across natural images, urban scenes and aerial imagery.
  }
  \par\vspace{-10pt}
  \label{fig:seg}
\end{figure}

\begin{wrapfigure}{r}{0.53\textwidth}
  \centering
  \includegraphics[width=\linewidth]{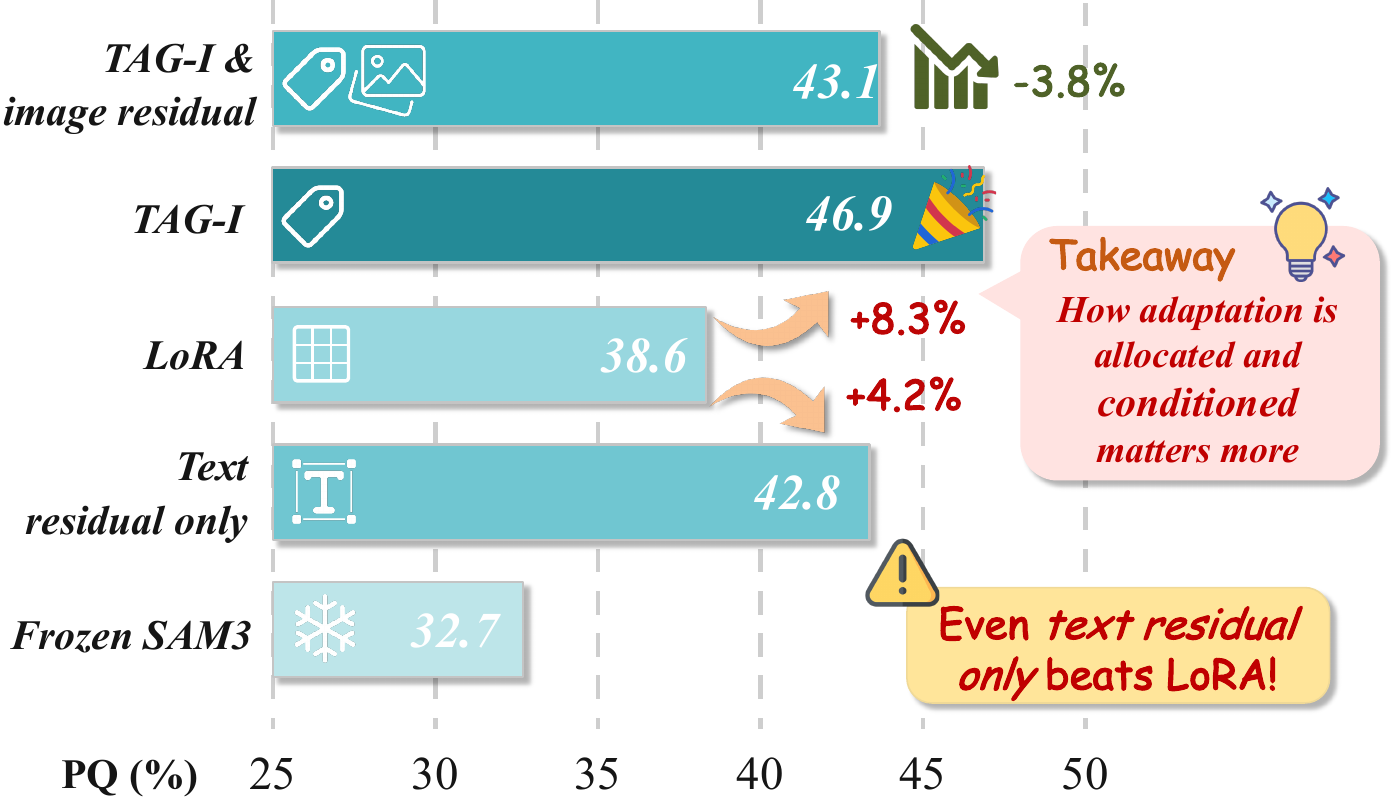}
  \setlength{\abovecaptionskip}{4pt}
  \vspace{-10pt}
  \caption{\textbf{Adaptation strategies on ADE20K.}
  PQ comparison of \TAGI{}, LoRA, and residual variants.}
  \label{fig:lora}
  \vspace{-4pt}
\end{wrapfigure}

\noindent\textbf{Targeted adaptation matters beyond low-rank capacity.}
To explore whether generic low-rank adaptation explains the gains of our work, we compare \TAGI{} against ordinary LoRA under identical supervision and objectives (see \textcolor{mylinkcolor}{Appendix~\ref{sec:appendix-targeted-adaptation}} for more details), as shown in  \textcolor{mylinkcolor}{Figure~\ref{fig:lora}}.
Although generic low-rank adaptation is already effective, text feature correction alone outperforms LoRA with a parameter budget matched to the full TAG by 4.2\% PQ on ADE20K, while completing the full ARP yields a further 4.1\% gain.
Besides, adding low-rank updates to the image features on top of TAG leads to a 3.8\% performance drop.
These results suggest that, under the current paradigm, the number of trainable parameters is not the key factor. 
What matters more is where adaptation is applied and how it is conditioned.
Guided by our empirical diagnostics and architectural insights into SAM3, ARP deliberately confines its intervention to category recognition and delivers the most substantial performance gains across all adaptation strategies.

\subsection{Qualitative Analysis}\label{sec:qualitative}

\textcolor{mylinkcolor}{Figure~\ref{fig:seg}} provides qualitative comparisons on three benchmarks.
Frozen SAM3 shows pronounced inter-class confusion and spurious activations, mislabeling objects and extending foreground predictions into background regions.
Through semantic anchor correction and presence calibration, \TAG{} alleviates these errors while retaining SAM3's spatial detail.
Both \TAGV{} and \TAGI{} yield more semantically consistent masks with cleaner object extents and less background leakage.
Further comparisons across diverse scene layouts and object scales appear in \textcolor{mylinkcolor}{Appendix~\ref{sec:appendix-qualitative}}.

\section{Conclusion}

In this paper, we present a systematic investigation into SAM3's performance breakdowns on universal image segmentation, pinpointing concept recognition as the governing bottleneck for downstream adaptation.
Motivated by this insight, we introduce \TAG{} to calibrate the text-conditioning process of SAM3, thereby unlocking the latent potential of its powerful segmentation capabilities.
We further design \textit{CWD} mechanism to effectively bridge high-level semantic context from MLLM with the precise spatial representations of SAM3, enabling semantic calibration in the unlabeled setting.
Empirical results demonstrate the efficiency and effectiveness of \TAG{}, which achieves impressive segmentation performance with only 3.1M trainable parameters.
Our work establishes a strong baseline for challenging universal image segmentation under weak supervision and label-free regimes, paving the way to match or exceed fully supervised SOTAs.


\clearpage
\appendix
\begingroup
\expandafter\let\csname endtabular*\endcsname\endtabular
\definecolor{apptableink}{HTML}{294C68}
\definecolor{apptablehead}{HTML}{E5EEF5}
\definecolor{apptablestripe}{HTML}{F3F6F8}
\definecolor{apptablegroup}{HTML}{EDF3F7}
\definecolor{apptablerule}{HTML}{718A9D}
\newcommand{\apptablestyle}[1]{%
  \centering\small
  \setlength{\tabcolsep}{#1}%
  \renewcommand{\arraystretch}{1.14}%
  \setlength{\extrarowheight}{0.5pt}%
  \setlength{\aboverulesep}{0pt}%
  \setlength{\belowrulesep}{0pt}%
  \setlength{\heavyrulewidth}{0.9pt}%
  \setlength{\lightrulewidth}{0.45pt}%
  \arrayrulecolor{apptablerule}%
}
\newcommand{\apphead}[1]{{\leavevmode\color{apptableink}\bfseries\boldmath #1}}
\newcommand{\appgroup}[1]{{\color{apptableink}\small\itshape #1}}

\raggedbottom
\setlength{\parskip}{3pt plus 0.5pt minus 0.5pt}
\setlength{\textfloatsep}{6pt plus 1pt minus 1pt}
\setlength{\floatsep}{6pt plus 1pt minus 1pt}
\setlength{\intextsep}{6pt plus 1pt minus 1pt}
\setlength{\abovecaptionskip}{4pt}
\setlength{\belowcaptionskip}{2pt}
\setlength{\abovedisplayskip}{7pt plus 1pt minus 1pt}
\setlength{\belowdisplayskip}{7pt plus 1pt minus 1pt}
\setlength{\columnsep}{12pt}
\renewcommand{\topfraction}{0.95}
\renewcommand{\bottomfraction}{0.92}
\renewcommand{\textfraction}{0.05}
\renewcommand{\floatpagefraction}{0.85}
\setcounter{topnumber}{3}
\setcounter{bottomnumber}{3}
\setcounter{totalnumber}{5}

\makeatletter
\setlength{\@fptop}{0pt}
\setlength{\@fpsep}{7pt plus 1pt minus 1pt}
\let\app@makecaption\@makecaption
\long\def\@makecaption#1#2{%
  \begingroup
  \ifdim\f@size pt>9pt\small\fi
  \app@makecaption{#1}{#2}%
  \endgroup}
\renewcommand{\@seccntformat}[1]{\csname the#1\endcsname.\quad}
\renewcommand{\section}{\@startsection{section}{1}{\z@}
  {-8pt plus -1pt minus -1pt}{4pt plus 1pt}
  {\normalfont\large\bfseries\raggedright}}
\renewcommand{\subsection}{\@startsection{subsection}{2}{\z@}
  {-7pt plus -1pt minus -1pt}{3pt plus 1pt}
  {\normalfont\normalsize\bfseries\raggedright}}
\renewcommand{\subsubsection}{\@startsection{subsubsection}{3}{\z@}
  {-6pt plus -1pt minus -1pt}{3pt plus 1pt}
  {\normalfont\normalsize\scshape\raggedright}}
\renewcommand{\paragraph}{\@startsection{paragraph}{4}{\z@}
  {4pt plus 1pt minus 1pt}{-1em}{\normalfont\normalsize\bfseries}}
\makeatother

\noindent The appendix is divided into several sections, each giving extra information and details.
\par

\startcontents[sections]
\printcontents[sections]{}{1}{%
  \setcounter{tocdepth}{2}%
  \fontsize{10}{12}\selectfont %
}
\par\addvspace{7pt}
\hrule
\addvspace{5pt}

\section{Related Work}\label{sec:related-work}

\noindent\textbf{Universal image segmentation} seeks a common framework for semantic~\citep{long2015fully}, instance~\citep{he2017mask}, and panoptic~\citep{kirillov2019panoptic} segmentation.
Building on DETR's set prediction~\citep{detr}, MaskFormer~\citep{maskformer} established mask classification as a unifying formulation, subsequently advanced through richer query--mask interaction~\citep{mask2former,zhang2021knet} and joint cross-task training~\citep{jain2023oneformer}.
MaskDINO~\citep{li2023maskdino} shares query representations between detection and segmentation, while OMG-Seg~\citep{li2024omg} extends task unification across image, video, open-vocabulary, and interactive segmentation.
TAPPS~\citep{degeus2024tapps} extends the structural granularity of panoptic segmentation by jointly representing objects and their parts with shared queries, explicitly linking each part prediction to its parent object.
Efficiency is another focus: PRO-SCALE~\citep{aich2025proscale} reduces encoder computation by progressively introducing finer-scale feature tokens at deeper layers, while RMP-SAM~\citep{xu2025rmpsam} uses dynamic convolution and decoupled prompt/object adapters for real-time interactive, panoptic, and video instance segmentation.
Other advances use a plain vision transformer~\citep{kerssies2025your} or decouple mask classification from mask generation~\citep{shahabodini2025missing}.
These advances broaden task coverage and improve efficiency under dense supervision.
Building on this progress, \TAG{} dispenses with dense annotations for downstream adaptation and uses only low-cost image-level labels to train lightweight adapters while keeping SAM3's pretrained weights frozen.
Reusing SAM3's pretrained spatial priors across all three tasks yields a favorable trade-off between annotation cost and segmentation performance.

\noindent\textbf{Open-vocabulary universal segmentation} extends task unification to categories specified by text at inference.
X-Decoder~\citep{zou2023xdecoder} integrates language with mask prediction, while OpenSeeD~\citep{zhang2023openseed} jointly learns from segmentation and detection data through a shared semantic space and decoding mechanisms that reconcile their annotation differences.
ODISE~\citep{xu2023odise} transfers representations from text-to-image diffusion, and FC-CLIP~\citep{yu2023convolutions} shares a frozen convolutional CLIP backbone between mask generation and recognition, avoiding separate image encoders for the two operations.
FreeSeg~\citep{qin2023freeseg} combines shared segmentation parameters with adaptive task and category prompts to support all three tasks in one model.
COSINE~\citep{liu2025cosine} unifies open-vocabulary and in-context segmentation through a shared decoder that aligns image features with textual and visual prompts.
HIPIE~\citep{wang2023hipie} further models whole objects, parts, and subparts hierarchically, with decoupled text--image fusion for thing and stuff categories.
These approaches establish flexible category interfaces using source-domain mask supervision.
Despite their ability to generalize to novel categories, direct transfer can fail under severe visual domain shift, as shown by our natural-to-aerial transfer results on iSAID (\textcolor{mylinkcolor}{Appendix~\ref{sec:appendix-ov-transfer}, Table~\ref{tab:ov-isaid-transfer}}).
Fine-tuning with masks from SIOR, an aerial domain closely related to iSAID, improves transfer, but obtaining dense annotations even in a related domain remains a barrier to real-world deployment.
\TAG{} requires no dense annotations for adaptation and achieves strong performance even on unlabeled target images by learning from MLLM-generated image-level labels (\textcolor{mylinkcolor}{Table~\ref{tab:universality}}).
This makes \TAG{} a more practical choice for closed-set applications with a predefined target vocabulary and limited annotation resources.

\noindent\textbf{Label-efficient universal segmentation} reduces the need for dense target annotations.
Progress in weakly supervised semantic~\citep{zhou2016learning,ru2023token} and instance~\citep{ahn2019weakly} segmentation has extended to broader task coverage with bounding boxes~\citep{li2018weakly,gu2023dataseg}, points in PSPS~\citep{fan2022pointly} and Point2Mask~\citep{li2023point2mask}, and image-level labels in JTSM~\citep{shen2021toward}.
These methods demonstrate the value of sparse supervision, while the limited spatial priors available during training can constrain the recovery of detailed boundaries and instance structure.
TRACE~\citep{jo2026trace} makes substantial progress by extracting instance boundaries from pretrained diffusion models and combining them with semantic predictions for tag-supervised panoptic segmentation.
Unsupervised approaches also advance mask discovery through self-supervised feature clustering~\citep{niu2024unsupervised} and divide-and-conquer grouping~\citep{wang2024unsam}.
\TAG{} contributes to this direction by reusing SAM3's segmentation priors and learning semantic calibration from image-level labels.
\textcolor{mylinkcolor}{Appendix Table~\ref{tab:weak-voc-coco}} compares these settings on VOC and COCO, including the published DHR+TRACE results.

\noindent\textbf{Segment Anything Models} have shifted the field toward large-scale promptable pretraining.
Built on the billion-scale SA-1B corpus, SAM~\citep{sam} translates geometric prompts into high-quality class-agnostic masks, and SAM2~\citep{sam2} carries the paradigm to video through streaming memory.
Subsequent variants sharpen boundaries~\citep{ke2023segment}, accelerate inference~\citep{zhang2023faster,xiong2024efficientsam}, or specialize to remote sensing~\citep{chen2024rsprompter}.
The class-agnostic formulation of SAM and SAM2 makes them useful spatial foundations for subsequent semantic integration.
Follow-up efforts couple them with grounded detectors~\citep{ren2024grounded}, multi-modal decoders~\citep{zou2023segment}, granularity-aware prompts~\citep{li2023semantic}, CLIP-based recognition~\citep{yuan2024open}, or language prompts encoded by an external vision-language model~\citep{openworldsam}.
SAM3~\citep{sam3} incorporates textual supervision during pretraining and uses a dedicated presence token to support concept recognition alongside localization.
\TAG{} builds on this text-conditioned pathway for target-domain calibration.

\noindent\textbf{Adapting SAM3 to target domains} is an emerging pursuit, as naive application often delivers suboptimal performance.
Mask-supervised fine-tuning~\cite{sam3} specializes SAM3 through pixel-level target annotations.
Among training-free methods, SegEarth-OV3~\cite{li2025segearthov3} and ActiveSAM~\cite{tien2026activesam} harness presence estimates to filter absent categories, while CoCo-SAM3~\cite{chen2026cocosam3} and ProC-SAM3~\cite{song2026prompt} enrich category prompts with linguistic variants and reconcile the ensuing predictions.
Straddling both routes, ConceptBank~\cite{conceptbank} distills target-calibrated class names from mask-annotated visual supports and LLM-expanded prompts.
These approaches improve concept prompting and prediction fusion.
\TAG{} learns corrections throughout the text-conditioned recognition pathway using image-level presence supervision.
Our diagnosis identifies substantial gains recoverable through category and activation correction with raw spatial predictions fixed, motivating this adaptation strategy.
Lightweight adapters then align text-conditioned interactions with the target domain while preserving all pretrained weights.
We directly compare \TAG{} with ConceptBank, CoCo-SAM3, ActiveSAM, and ProC-SAM3 in \textcolor{mylinkcolor}{Appendix~\ref{sec:appendix-sam3-calibration} (Table~\ref{tab:sam3-calibration})}.

\section{Counterparts}\label{sec:appendix-counterparts}
This section describes the counterparts in \textcolor{mylinkcolor}{Table~\ref{tab:universality}}, their supervision settings, and the sources of the reported results.
We distinguish training with target-domain masks from transfer using source-domain masks, and specify how each setting is evaluated on aerial imagery.

\subsection{Fully Supervised Universal Segmentation}
We compare with Mask2Former~\citep{mask2former}, MaskDINO~\citep{li2023maskdino}, OneFormer~\citep{jain2023oneformer}, PEM~\citep{cavagnero2024pem}, EoMT~\citep{kerssies2025your}, and ViT-P~\citep{shahabodini2025missing}.
These methods advance universal segmentation through query-based mask prediction, joint task modeling, efficient decoding, and pretrained vision transformers.
In the fully supervised setting, a model is trained with ground-truth masks from the target domain and evaluated on held-out images from that same domain.
These results therefore provide references with stronger supervision than \TAG{}.

Among our three evaluation benchmarks, the original papers primarily report results on ADE20K and Cityscapes.
For these two datasets, \textcolor{mylinkcolor}{Table~\ref{tab:universality}} quotes the published scores, retaining the reported task-specific configurations and leaving unreported metrics blank.
To extend the comparison to aerial imagery, we reproduce Mask2Former, MaskDINO, OneFormer, PEM, and EoMT using their official repositories, train them with iSAID training masks, and evaluate them on the iSAID validation set.
ViT-P is included as a published reference on ADE20K and Cityscapes; no iSAID reproduction is reported.

\subsection{Open-Vocabulary Universal Segmentors}\label{sec:appendix-ov-transfer}
We compare with X-Decoder~\citep{zou2023xdecoder}, HIPIE~\citep{wang2023hipie}, APE~\citep{shen2024ape}, SAM-CP~\citep{chen2025samcp}, OpenWorldSAM~\citep{openworldsam}, and FrozenSeg~\citep{chen2024frozenseg}.
These methods learn segmentation from source-domain masks and use language-conditioned recognition to predict categories beyond the source vocabulary, without requiring target-domain masks.
A common protocol trains on COCO panoptic annotations and evaluates on other natural-image benchmarks, including ADE20K and Cityscapes; some methods additionally use detection, grounding, or image--text data.
We quote the published natural-image results in \textcolor{mylinkcolor}{Table~\ref{tab:universality}} and evaluate the released pretrained models on aerial imagery.

\paragraph{Direct transfer under domain shift.}
Transfer from natural images to aerial imagery introduces a much larger change in viewpoint, object scale, and scene appearance.
\textcolor{mylinkcolor}{Table~\ref{tab:ov-isaid-transfer}} evaluates direct transfer to iSAID using the full target vocabulary.
Across the five evaluated methods, direct-transfer scores range from 1.32 to 6.05 PQ, 0.04 to 3.09 mask AP, and 3.56 to 8.08 mIoU.
These measurements characterize the challenge of transferring natural-image representations to overhead scenes and motivate examining adaptation with related-domain imagery.

\begin{table}[!htbp]
\apptablestyle{4pt}
\caption{\textbf{Open-vocabulary transfer to iSAID before and after SIOR adaptation.}
Direct transfer uses models pretrained on natural-image data; adaptation uses SIOR masks, without iSAID training masks.
Both settings evaluate the full iSAID vocabulary on 11,644 validation tiles; per-image GT class lists are never supplied.
The adapted scores correspond to the iSAID open-vocabulary entries in \textcolor{mylinkcolor}{Table~\ref{tab:universality}}.
AP denotes mask AP. Bold marks the best score within each setting.}
\label{tab:ov-isaid-transfer}
\resizebox{\columnwidth}{!}{%
\begin{tabular}{ll!{\vrule width 0.6pt}rrr!{\vrule width 0.4pt}rrr}
\toprule
\rowcolor{apptablehead}
& & \multicolumn{3}{c!{\vrule width 0.4pt}}{\apphead{Direct transfer}} & \multicolumn{3}{>{\columncolor{oursrow}}c}{\apphead{After SIOR adaptation}} \\
\rowcolor{apptablehead}
\apphead{Method} & \apphead{Backbone} & \apphead{PQ} & \apphead{AP} & \apphead{mIoU} & \cellcolor{oursrow}\apphead{PQ} & \cellcolor{oursrow}\apphead{AP} & \cellcolor{oursrow}\apphead{mIoU} \\
\midrule
\rowcolor{white}
X-Decoder~\citep{zou2023xdecoder} & DaViT-L & 1.32 & 0.33 & 4.78 & 28.09 & 17.61 & 37.10 \\
\rowcolor{apptablestripe}
HIPIE~\citep{wang2023hipie} & ViT-H & 2.28 & 0.57 & 5.53 & 28.56 & 13.17 & 9.53 \\
\rowcolor{white}
APE~\citep{shen2024ape} & EVA-02 ViT-L & \textbf{6.05} & \textbf{3.09} & 7.62 & 32.86 & \textbf{23.54} & \textbf{37.36} \\
\rowcolor{apptablestripe}
OpenWorldSAM~\citep{openworldsam} & SAM2-L + BEiT-3 & 2.22 & 0.04 & 3.56 & 21.6 & 10.5 & 22.3 \\
\rowcolor{white}
FrozenSeg~\citep{chen2024frozenseg} & ConvNeXt-L + SAM & 3.26 & 2.97 & \textbf{8.08} & \textbf{33.54} & 19.18 & 32.53 \\
\bottomrule
\end{tabular}}
\end{table}

\paragraph{SIOR-to-iSAID evaluation.}
To provide a source domain with more closely related visual characteristics, we adapt the methods on SIOR~\citep{wang2023samrs}, a separate aerial dataset, using its dense mask annotations and each method's official implementation.
We then evaluate on iSAID with the full target vocabulary and no iSAID training masks.
This preserves the source-to-target transfer protocol while providing mask supervision from a domain closer to the test images.
The improvements in \textcolor{mylinkcolor}{Table~\ref{tab:ov-isaid-transfer}} demonstrate the benefit of this related-domain supervision; for example, OpenWorldSAM improves from 2.22 to 21.6 PQ after SIOR adaptation.
Accordingly, the iSAID open-vocabulary entries in \textcolor{mylinkcolor}{Table~\ref{tab:universality}} report these SIOR-adapted results, and their trainable parameter counts refer to SIOR fine-tuning.
The \supMs{} notation distinguishes this source-mask supervision from \TAG{}'s target image-level supervision.

\subsection{Weakly Supervised Methods}
JTSM~\citep{shen2021toward}, PSPS~\citep{fan2022pointly}, and Point2Mask~\citep{li2023point2mask} establish important steps toward universal segmentation with image-level or point annotations, primarily on VOC and COCO.
Point2Mask uses semantic and boundary cues within an optimal-transport formulation to derive dense supervision from sparse points.
Although these approaches have advanced weakly supervised segmentation, learning detailed spatial structure remains challenging with limited spatial priors and no pretrained foundation-model mask decoder.
Our Point2Mask reproductions with an R50 backbone cover ADE20K, Cityscapes, and iSAID, extending the evaluation across natural images, urban scenes, and aerial imagery (\textcolor{mylinkcolor}{Table~\ref{tab:universality}}).
On ADE20K, Point2Mask achieves 18.6 PQ, 11.8 mask AP, and 26.3 mIoU under point supervision.
The comparison motivates studying how stronger pretrained spatial priors can complement weak semantic supervision.
TRACE~\citep{jo2026trace} makes substantial progress in this direction by extracting instance boundaries from diffusion-model representations.
Combined with DHR semantics, it achieves strong tag-supervised panoptic results on VOC and COCO.
\textcolor{mylinkcolor}{Table~\ref{tab:weak-voc-coco}} provides a direct comparison with the published DHR+TRACE results on VOC and COCO, alongside other weakly supervised methods including EPLD~\citep{li2024learnable}.
This comparison uses the stated panoptic benchmarks and retains each method's reported supervision and backbone.
\textcolor{mylinkcolor}{Appendix~\ref{sec:appendix-weak}} details the protocols and the gains achieved by \TAG{}.

\section{Detailed Experimental Setup}\label{sec:appendix-setup}\label{sec:appendix-method}
This section specifies the implementation and hyperparameters of \TAG{} and the architectures used in the targeted-adaptation comparison.

\subsection{Implementation Details}\label{sec:appendix-training}

\paragraph{Shared training configuration.}
We use the same training hyperparameters on all datasets without dataset-specific tuning, as summarized in \textcolor{mylinkcolor}{Table~\ref{tab:tag-training}}.
All models are trained for 10 epochs with a peak learning rate of $5{\times}10^{-3}$, presence weight $\lambda_{\mathrm{pres}}=4$, ranking weight $\lambda_{\mathrm{rank}}=0.5$, hard-negative count $m=3$, adapter rank $r=16$, and $G=16$ shared presence directions.
With this common configuration, \TAG{} consistently improves on frozen SAM3 across the evaluated domains in \textcolor{mylinkcolor}{Table~\ref{tab:universality}}.
These results support the generality of the adaptation strategy across natural, urban, and aerial imagery without per-dataset hyperparameter optimization.
\textcolor{mylinkcolor}{Appendix~\ref{sec:appendix-hyperparameter-sensitivity}} further examines sensitivity to objective weights, negative mining, and adapter capacity.

\begin{table}[!htbp]
\apptablestyle{8pt}
\caption{\textbf{Shared training hyperparameters across datasets.}
The same configuration is used without dataset-specific tuning.
The loss weights follow \textcolor{mylinkcolor}{Eq.~\ref{eq:loss}}, with CWD replacing BCE for \TAGV{}.
Epochs denote the scheduled training budget.}
\label{tab:tag-training}
\begin{tabularx}{0.72\linewidth}{l>{\centering\arraybackslash}X}
\toprule
\rowcolor{apptablehead}
\apphead{Setting} & \apphead{Shared value} \\
\midrule
\rowcolor{white}
Learning rate & $5{\times}10^{-3}$ \\
\rowcolor{apptablestripe}
Training epochs & 10 \\
\midrule
\rowcolor{white}
Presence weight $\lambda_{\mathrm{pres}}$ & 4 \\
\rowcolor{apptablestripe}
Ranking weight $\lambda_{\mathrm{rank}}$ & 0.5 \\
\rowcolor{white}
Hard negatives $m$ & 3 \\
\rowcolor{apptablestripe}
Adapter rank $r$ & 16 \\
\rowcolor{white}
Presence directions $G$ & 16 \\
\bottomrule
\end{tabularx}
\end{table}

\paragraph{Repeated runs and stability.}
For each dataset, the \TAGI{} and \TAGV{} scores in \textcolor{mylinkcolor}{Table~\ref{tab:universality}} are arithmetic means over three independent training runs with different random seeds and otherwise identical settings.
We observe little variation across runs, with a maximum sample variance of 0.3 squared percentage points across the reported dataset--metric combinations.
These small fluctuations indicate stable performance across the evaluated seeds.

\paragraph{Model and optimization.}
We implement \TAG{} in PyTorch with the pretrained image branch of SAM3~\citep{sam3} and conduct experiments on NVIDIA RTX 4090 GPUs.
The image and text encoders, fusion encoder, transformer decoder, and prediction heads retain their pretrained weights; only ARP parameters are optimized.
We use Adam~\citep{kingma2014adam} with weight decay $10^{-4}$, numerical stabilizer $10^{-8}$, and gradient-norm clipping at $0.1$.
The cosine OneCycle schedule~\citep{smith2019super} warms up for one epoch, starting at $1/25$ of the learning rate in \textcolor{mylinkcolor}{Table~\ref{tab:tag-training}} and ending at $1/10^{4}$ of that initial rate.
Its first-moment coefficient cycles between $0.85$ and $0.95$, while the second-moment coefficient is fixed at $0.999$.
The learning rates of the presence, text-feature, fusion K/V, and decoder K/V parameter groups are scaled by $0.1$, $0.1$, $1.0$, and $0.5$, respectively.
The forward computation uses bfloat16 autocasting, with loss evaluation in float32.

\paragraph{Loss settings.}
For \TAGI{}, the training objective combines BCE and SBR using the shared weights in \textcolor{mylinkcolor}{Table~\ref{tab:tag-training}}.
\TAGV{} replaces BCE with CWD and retains the same loss weights on every dataset.
SBR selects up to $m$ highest-scoring negatives among the supervised classes.
Selection uses detached logits, while gradients flow through the selected positive and negative logits.
The ranking term is zero when either set is empty; BCE contributes zero when no valid classes remain, and CWD contributes zero when the total confidence weight is zero.
An explicit background category is excluded from presence supervision.

\paragraph{MLLM supervision and shared CWD settings.}\label{sec:appendix-cwd}
Qwen3.8-27B generates category-presence scores offline using the prompt described in \textcolor{mylinkcolor}{Sec.~\ref{sec:vlm}}.
Condensed prompt templates and example answers with presence probabilities for ADE20K, Cityscapes, and iSAID are provided in \textcolor{mylinkcolor}{Appendix~\ref{sec:appendix-prompts}}.
The scores are cached for training, so no MLLM is required at segmentation inference.
We also keep all CWD hyperparameters fixed across datasets: $h=0.6$, $\ell=0.1$, $\epsilon_{\mathrm{pos}}=0.05$, and $\epsilon_{\mathrm{neg}}=0.1$.
These choices are guided primarily by prior knowledge: a relatively high $h$ favors confident positive labels, while a low $\ell$ selects reliable negative classes for SBR.
Small smoothing rates soften the binary targets while preserving the distinction between positive and negative supervision.
With these shared settings, \TAGV{} improves all nine main-benchmark metrics over frozen SAM3, demonstrating the effectiveness of a common supervision recipe across domains.
The corresponding distance scale in \textcolor{mylinkcolor}{Eq.~\ref{eq:cwd}} is $\beta=(h-\ell)/2=0.25$.
All reported CWD runs use the linear, clipped confidence weight centered at $h$.
Only classes with $p_k\le\ell$ or $p_k\ge h$ enter SBR; classes within the uncertainty band can still contribute to the confidence-weighted BCE.

\paragraph{Image preparation.}
We use canonical class names as text prompts and reuse their frozen encodings across images.
Image features are computed once per input view and shared across class prompts.
SAM3 resizes each view to $1008{\times}1008$ and normalizes RGB channels with mean and standard deviation both equal to $0.5$.

\paragraph{Pooling and initialization.}
Let $\mu_{tk}\in\{0,1\}$ indicate whether token $t$ of class $k$ is valid.
The pooled anchor used in \textcolor{mylinkcolor}{Eq.~\ref{eq:template}} is
\begin{equation}\label{eq:pool}
  \bar{\mathbf e}_k
  =\frac{\sum_{t=1}^{N_t}\mu_{tk}\,(\mathbf F_t^k)_{t,:}}
  {\max\!\left(\sum_{t=1}^{N_t}\mu_{tk},1\right)}.
\end{equation}
Each token row $(\mathbf F_t^k)_{t,:}$ lies in $\mathbb R^{1\times d}$.
Applying the same mask to $\mathbf F_t^{\prime k}(I)$ yields $\bar{\mathbf e}'_k$, which conditions the presence and K/V adapters.
For ARP, low-rank input projections and the shared presence directions are initialized from a zero-mean Gaussian with standard deviation $10^{-3}$, and the image-affinity scale $\eta$ starts at one.
\textcolor{mylinkcolor}{Table~\ref{tab:arp}} lists the zero-initialized output projections, which make every residual vanish at initialization.
The gate's final layer also starts at zero, giving a multiplicative gate of one in \textcolor{mylinkcolor}{Eq.~\ref{eq:template}}.

\begin{table}[!htbp]
\apptablestyle{4pt}
\caption{\textbf{ARP inputs and zero initialization.}
Each residual starts at zero through its output projection.
K/V updates are applied before splitting features into attention heads.}
\label{tab:arp}
\begin{tabularx}{\linewidth}{ll>{\raggedright\arraybackslash}p{0.25\linewidth}>{\raggedright\arraybackslash}X}
\toprule
\rowcolor{apptablehead}
\apphead{Module} & \apphead{Residual input} & \apphead{Conditioning feature} & \apphead{Zero-initialized component} \\
\midrule
\rowcolor{white}
Text low-rank branch & $\mathbf F_t^k$ & $\bar{\mathbf e}_k$ & output projection $\mathbf B$ \\
\rowcolor{apptablestripe}
Text image branch & $\mathbf F_t^k,\mathbf F_v$ & image features and text affinity & $\operatorname{Proj}_{\mathrm{out}}$ weight/bias \\
\rowcolor{white}
Presence token & $\bar{\mathbf e}'_k$ & $\bar{\mathbf e}'_k$ & final layer of $\mathrm{MLP}_{\mathrm p}$ \\
\rowcolor{apptablestripe}
Fusion text K/V & $\mathbf F_t^{\prime k}(I)$ & $\bar{\mathbf e}'_k$ & both low-rank output projections \\
\rowcolor{white}
Decoder image K/V & $\mathbf F_v^k$ & $\bar{\mathbf e}'_k$ & both low-rank output projections \\
\bottomrule
\end{tabularx}
\end{table}

\subsection{Targeted Adaptation Architectures}
\label{sec:appendix-targeted-adaptation}

\textcolor{mylinkcolor}{Figure~\ref{fig:lora}} compares alternative adaptation architectures for ADE20K panoptic segmentation.
All trainable variants use image-level category-presence labels and the same BCE and SBR objectives.
The pretrained SAM3 weights remain frozen, and only the parameters introduced by each adaptation strategy are optimized.
Frozen SAM3 provides the unadapted reference.

\paragraph{Ordinary LoRA.}
We replace ARP with conventional LoRA~\citep{lora} at two sets of cross-attention projections.
In the fusion encoder, LoRA adapts the \emph{textual key and value projections} used by image-to-text cross-attention.
In the decoder, it adapts the \emph{visual key and value projections} used by query-to-image cross-attention.

\begin{samepage}
For one such projection, let $\mathbf X$ denote its input tokens and $\mathbf W\in\mathbb R^{d_{\mathrm{in}}\times d_{\mathrm{out}}}$ its frozen pretrained weight, where $d_{\mathrm{in}}$ and $d_{\mathrm{out}}$ are the input and output widths.
We introduce trainable factors $\mathbf A_{\mathrm L}\in\mathbb R^{d_{\mathrm{in}}\times r_{\mathrm L}}$ and $\mathbf B_{\mathrm L}\in\mathbb R^{r_{\mathrm L}\times d_{\mathrm{out}}}$, where $r_{\mathrm L}$ is the LoRA rank.
The adapted projection is
\begin{equation}\label{eq:lora-projection}
  \mathbf X\mathbf W
  +(\mathbf X\mathbf A_{\mathrm L})\mathbf B_{\mathrm L}.
\end{equation}
Only these factors are optimized; the original projection weights and biases remain frozen.
The factors are shared across categories and applied without an explicit class-conditioned gate.
\end{samepage}

No low-rank updates are added to self-attention, query projections, or attention-output projections.
The image and text encoders, feed-forward and normalization layers, prediction heads, and pretrained presence-token embedding also remain frozen.
The baseline omits TAG's text-feature adapter, image-conditioned text residual, and class-conditioned presence-token initialization module.
The rank allocation is chosen to give a trainable parameter budget comparable to that of the complete \TAGI{} model.

\paragraph{Text residual only.}
This variant confines the added adaptation modules to the text-feature correction in \textcolor{mylinkcolor}{Eq.~\ref{eq:text-residual}}.
It retains both the class-conditioned static residual and the image-conditioned text-to-image cross-attention branch, allowing the text anchors to incorporate target-domain semantics and image-specific evidence.
The corrected text tokens then enter SAM3's pretrained conditioning pathway.
The variant omits the additional fusion and decoder K/V residuals and the class-conditioned presence-token initialization.

\paragraph{Complete \TAGI{}.}
The full model retains this text-feature correction and adds the remaining ARP components described in \textcolor{mylinkcolor}{\S\ref{sec:arp}}.
These components adapt the fusion encoder's text keys and values, initialize the presence token from the corrected class anchor, and condition the decoder's image K/V residuals on that anchor.

\paragraph{\TAGI{} \& image residual.}
This variant retains the complete ARP and adds a residual branch directly to the image tokens produced by the frozen image encoder and its feature neck, before text-conditioned fusion.
For the visual tokens $\mathbf F_v$, a learned low-rank mapping $\Delta_{\mathrm{vis}}$ produces corrected image tokens
\begin{equation}\label{eq:visual-token-residual}
  \widetilde{\mathbf F}_v
  =\mathbf F_v+\Delta_{\mathrm{vis}}(\mathbf F_v),
\end{equation}
where both $\widetilde{\mathbf F}_v$ and $\Delta_{\mathrm{vis}}(\mathbf F_v)$ have the same shape as $\mathbf F_v$.
The mapping is shared across spatial positions and class prompts, and its output projection is initialized to zero.
The corrected visual representation is supplied to both the image-conditioned text branch and the fusion encoder, whose output subsequently feeds the decoder.
The added branch is trained jointly with ARP under the same presence objective.
In \textcolor{mylinkcolor}{Eq.~\ref{eq:image-residual}}, $\Delta_{\mathrm{img}}^k(I)$ uses visual evidence to update \emph{text tokens}; here, $\Delta_{\mathrm{vis}}$ directly updates \emph{image tokens}.

\Needspace{4\baselineskip}
\section{More Quantitative Experiments}\label{sec:appendix-quantitative}
We extend the quantitative evaluation with additional ablation studies, comparisons on VOC and COCO, direct comparisons with SAM3 calibration methods, a comparison of learned conditioning and retrieved class descriptions, object-level recognition and spatial diagnostics, and vocabulary-expansion results on ADE847.

\begingroup
\raggedbottom
\subsection{Ablation Studies}\label{sec:appendix-ablation}
We extend the ablations in \textcolor{mylinkcolor}{Sec.~\ref{sec:ablation}} to examine the internal conditioning pathway, the choice of MLLM supervision, and the sensitivity to objective weights and adapter capacity.

\Needspace{4\baselineskip}
\subsubsection{Module Design}
\begin{figure}[!htbp]
  \centering
  \includegraphics[width=0.95\linewidth,trim=24bp 14bp 20bp 14bp,clip]{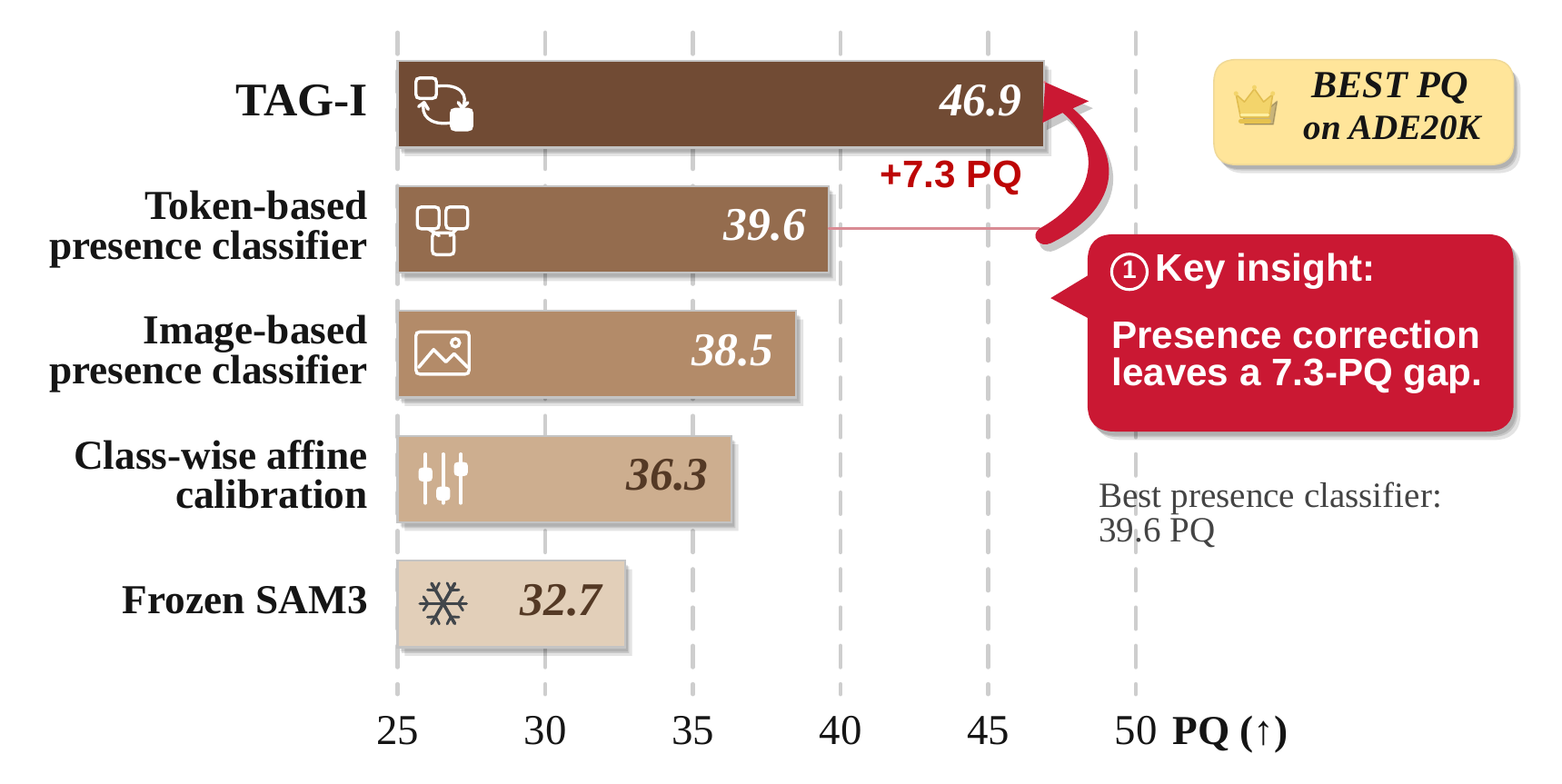}
  \par\vspace{-4pt}
  \setlength{\abovecaptionskip}{2pt}
  \caption{\textbf{Presence correction on ADE20K.}
  PQ of score calibration, presence classifiers, and \TAGI{}; the axis starts at 25.}
  \label{fig:appendix-readout-controls}
\end{figure}

\paragraph{Beyond presence-score correction.}
\textcolor{mylinkcolor}{Figure~\ref{fig:appendix-readout-controls}} compares score calibration and presence classifiers with \TAGI{} on ADE20K.
All pretrained SAM3 weights remain frozen; only the added modules are trained.

\noindent\textbf{Frozen SAM3.}
The original model is used without additional modules or adaptation.

\noindent\textbf{Class-wise affine calibration.}
Only a per-category scale and offset are learned to recalibrate the original presence logits.

\noindent\textbf{Image-based presence classifier.}
A small classifier corrects the presence logits using pooled features from the frozen image encoder.
Only this classifier is trained.

\noindent\textbf{Token-based presence classifier.}
A small classifier combines the frozen decoder's presence-token features with the queried class's text features to correct the presence logits.
Only the classifier is trained; token initialization and the decoder remain unchanged.

\noindent\textbf{\TAGI{}.}
ARP corrects text features and adds class-conditioned residuals to the fusion encoder, decoder, and presence-token initialization.

The three output-level controls preserve SAM3's raw mask predictions and change only the presence scores used to filter and combine them.

All three controls improve PQ over frozen SAM3.
The token-based classifier reaches 39.6 PQ, the highest score among these controls, and adapting the conditioning pathway with \TAGI{} provides a further 7.3 points.
These results show that correcting presence scores on fixed features is useful, while adapting how text features guide segmentation provides further gains.

\par
\Needspace{4\baselineskip}
\subsubsection{Text Correction and MLLM Supervision}\label{sec:appendix-adapter-mllm}
\noindent\textbf{Complementary text and image-conditioned corrections.}\par\nobreak
\begin{table}[!htbp]
  \apptablestyle{4pt}
  \setlength{\abovecaptionskip}{0pt}
  \setlength{\belowcaptionskip}{5pt}
  \caption{\textbf{Complementary text corrections.}
  Branch ablations of \TAGI{}; PQ (\%).}
  \label{tab:appendix-text-branches}
  \begin{tabularx}{\linewidth}{@{}p{0.32\linewidth}*{5}{>{\centering\arraybackslash}X}@{}}
    \toprule
    \rowcolor{apptablehead}
    \apphead{Method} & \apphead{ADE} & \apphead{City} & \apphead{iSAID} & \apphead{Avg.} & \apphead{$\Delta$} \\
    \midrule
    \rowcolor{white}
    w/o text low-rank & 43.51 & 59.73 & 41.39 & 48.21 & $-3.05$ \\
    \rowcolor{apptablestripe}
    w/o image conditioning & 44.19 & 61.02 & 42.60 & 49.27 & $-1.99$ \\
    \rowcolor{apptablegroup}
    \TAGI{} & \textbf{46.86} & \textbf{62.67} & \textbf{44.24} & \textbf{51.26} & -- \\
    \bottomrule
  \end{tabularx}
  \par\vspace{4pt}
  \raggedright\fontsize{8}{9.5}\selectfont
  ADE: ADE20K; City: Cityscapes. Avg.: mean over the three datasets; $\Delta$: difference from full \TAGI{} in PQ points.
\end{table}

We ablate the two branches that update the text features in \textcolor{mylinkcolor}{Eq.~\ref{eq:text-residual}}.
As shown in \textcolor{mylinkcolor}{Table~\ref{tab:appendix-text-branches}}, removing the text low-rank branch reduces average PQ from 51.26 to 48.21, while removing image conditioning lowers it to 49.27.
Retaining both branches gives higher PQ on all three datasets, supporting their complementary roles.
The image-invariant text correction learns category-dependent adjustments that account for the shift from pretrained semantics to the target-domain distribution.
The image-conditioned branch injects visual evidence from the current image into the text anchors, making their correction responsive to the observed scene.
Together, they combine target-domain semantic correction with image-specific grounding.
\par

\Needspace{4\baselineskip}
\noindent\textbf{MLLM choice and confidence-weighted distillation.}\par\nobreak
\begin{figure}[!htbp]
  \centering
  \includegraphics[width=0.70\linewidth,trim=1bp 5bp 0.5bp 2bp,clip]{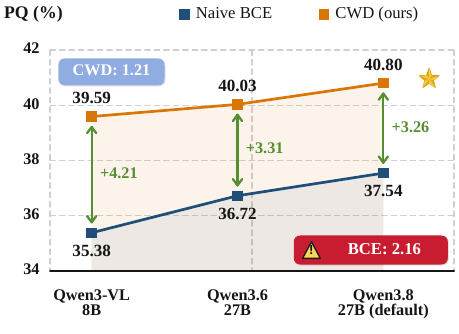}
  \par\vspace{-4pt}
  \setlength{\abovecaptionskip}{2pt}
  \caption{\textbf{MLLM supervision on ADE20K.}
  Naive BCE versus CWD; PQ (\%). Green arrows show CWD gains; spans are the ranges across MLLMs.}
  \label{fig:appendix-adapter-mllm}
\end{figure}
We focus on open-source MLLMs for offline image-level tagging, avoiding paid API calls.
\textcolor{mylinkcolor}{Figure~\ref{fig:appendix-adapter-mllm}} compares naive BCE with CWD on ADE20K.
CWD improves PQ by 4.21, 3.31, and 3.26 points with Qwen3-VL 8B, Qwen3.6 27B, and the default Qwen3.8 27B, respectively.
Its cross-MLLM performance range is 1.21 points, compared with 2.16 for BCE, indicating lower sensitivity to the tag generator.
CWD downweights decisions near the confidence threshold and smooths targets; the accompanying ranking term also excludes uncertain classes.
The consistent gains across all three teachers demonstrate that CWD makes effective use of automatically generated presence labels.
The default Qwen3.8 27B supervision gives the highest PQ among the tested choices, supporting its use for adapting \TAG{} on unlabeled target images.
\par

\Needspace{4\baselineskip}
\subsubsection{Hyperparameter Sensitivity}\label{sec:appendix-hyperparameter-sensitivity}
\paragraph{A shared configuration across domains.}
All main-benchmark \TAG{} results use the same training configuration in \textcolor{mylinkcolor}{Table~\ref{tab:tag-training}}, without dataset-specific tuning.
For \TAGI{}, the common setting $(\lambda_{\mathrm{pres}},\lambda_{\mathrm{rank}},m,r,G)=(4,0.5,3,16,16)$ achieves 46.9, 62.7, and 44.2 PQ on ADE20K, Cityscapes, and iSAID, respectively, as reported in \textcolor{mylinkcolor}{Table~\ref{tab:universality}}.
The consistent gains over frozen SAM3 across these different domains support the generality of \TAG{} under a shared training recipe.

\paragraph{Sensitivity protocol.}
To assess sensitivity to individual design choices, we vary the presence weight $\lambda_{\mathrm{pres}}$, ranking weight $\lambda_{\mathrm{rank}}$, hard-negative count $m$, adapter rank $r$, and number of shared presence directions $G$ one at a time.
All remaining parameters retain the shared configuration above, including the learning rate of $5{\times}10^{-3}$ and the training budget of 10 epochs.
\textcolor{mylinkcolor}{Figures~\ref{fig:appendix-loss-sensitivity} and~\ref{fig:appendix-structure-sensitivity}} report absolute PQ on the three datasets, with the common default highlighted in each sweep.

\begin{figure}[!htbp]
\centering
\includegraphics[width=\linewidth]{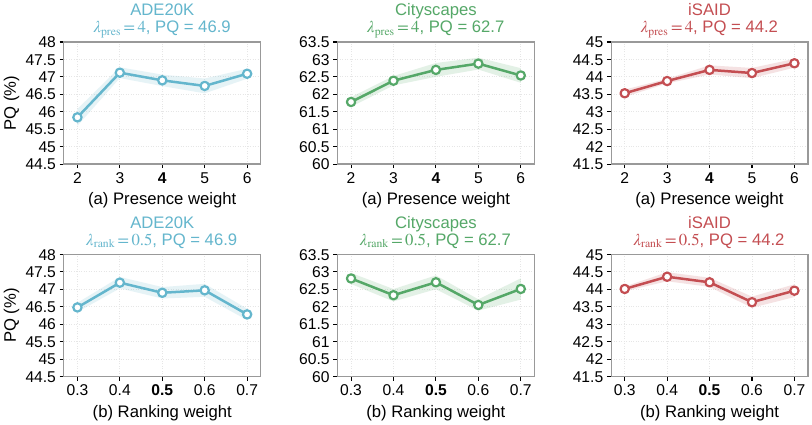}
\caption{\textbf{Sensitivity to objective weights.}
Rows vary (a) the presence weight and (b) the ranking weight while fixing all other settings to \textcolor{mylinkcolor}{Table~\ref{tab:tag-training}}.
Columns show ADE20K, Cityscapes, and iSAID on separate absolute-PQ axes, each spanning 3.5 points.
Axis limits are fixed for each dataset across all sweeps.
Colored subtitles report the common setting and its PQ; bold x-axis ticks identify that setting.}
\label{fig:appendix-loss-sensitivity}
\end{figure}

\paragraph{Stable performance across objective weights.}
\textcolor{mylinkcolor}{Figure~\ref{fig:appendix-loss-sensitivity}} varies $\lambda_{\mathrm{pres}}$ from 2 to 6 and $\lambda_{\mathrm{rank}}$ from $0.3$ to $0.7$.
The largest within-dataset PQ spans are 1.28 and 0.91 points, respectively, and the common defaults remain within 0.29 PQ of every sweep's best result.
The shared $\lambda_{\mathrm{pres}}=4$ setting improves PQ over $\lambda_{\mathrm{pres}}=2$ by 1.06, 0.92, and 0.67 points on ADE20K, Cityscapes, and iSAID, respectively.
This comparison supports assigning sufficient weight to presence calibration alongside SBR in \textcolor{mylinkcolor}{Eq.~\ref{eq:loss}}.
The small default-to-best gaps show that the shared loss weights provide consistently strong performance throughout the evaluated domains.

\begin{figure}[!htbp]
\centering
\includegraphics[width=\linewidth]{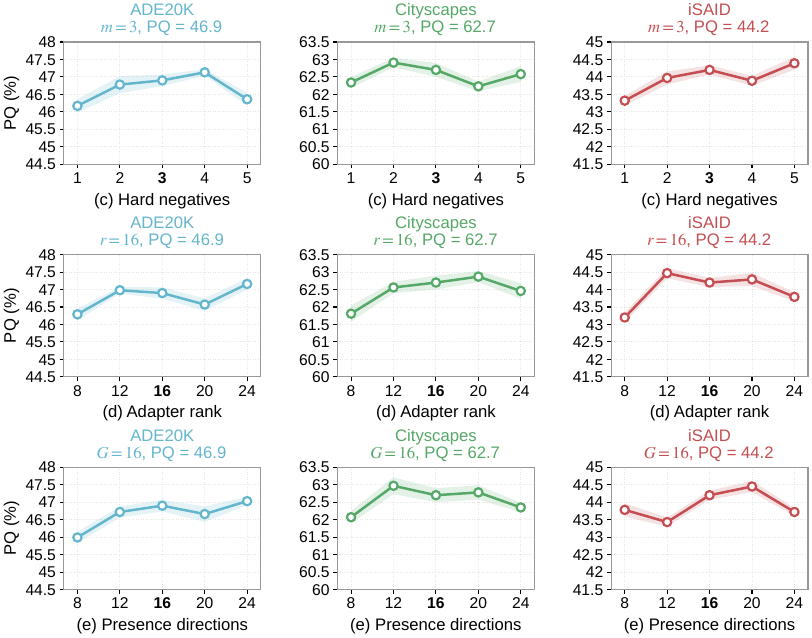}
\caption{\textbf{Sensitivity to hard-negative mining and adapter capacity.}
Rows show one-factor sweeps of (c) hard-negative count $m$, (d) adapter rank $r$, and (e) shared presence directions $G$.
The common defaults are $m=3$, $r=16$, and $G=16$, with $\lambda_{\mathrm{pres}}=4$ and $\lambda_{\mathrm{rank}}=0.5$ fixed.
Dataset columns, colors, default annotations, and the 3.5-point axis spans follow \textcolor{mylinkcolor}{Figure~\ref{fig:appendix-loss-sensitivity}}.}
\label{fig:appendix-structure-sensitivity}
\end{figure}

\paragraph{Stable performance across mining and capacity choices.}
\textcolor{mylinkcolor}{Figure~\ref{fig:appendix-structure-sensitivity}} varies $m$ from 1 to 5 and both $r$ and $G$ from 8 to 24.
The maximum PQ spans are 1.07, 1.27, and 1.04 points for $m$, $r$, and $G$, respectively.
The default $r=16$ improves PQ over $r=8$ by 0.61, 0.89, and 1.00 points on ADE20K, Cityscapes, and iSAID, respectively.
Across all mining and capacity sweeps, the common defaults remain within 0.27 PQ of the best result.
This combination of compact adapter capacity and near-best accuracy supports the shared choices of $m=3$, $r=16$, and $G=16$.

\paragraph{Shared settings across domains.}
Across all fifteen dataset--parameter pairs, the maximum--minimum difference in central PQ ranges from 0.68 to 1.28 points, while the shared configuration remains within 0.29 PQ of every sweep's best result.
These bounded variations support a stable operating range for \TAGI{} under individual hyperparameter changes.
We therefore use $(4,0.5,3,16,16)$ on every dataset and keep the CWD settings fixed for \TAGV{}, as specified in \textcolor{mylinkcolor}{Appendix~\ref{sec:appendix-training}}.
The resulting gains of 14.2, 12.4, and 25.0 PQ points over frozen SAM3 on ADE20K, Cityscapes, and iSAID demonstrate effective adaptation across natural images, urban scenes, and aerial imagery with one training recipe.
\endgroup

\Needspace{4\baselineskip}
\subsection{Experiments on VOC and COCO}\label{sec:appendix-weak}
\subsubsection{Benchmarks and Metric Conventions}
We complement the three challenging benchmarks in \textcolor{mylinkcolor}{Table~\ref{tab:universality}} with the VOC and COCO settings commonly used in weakly supervised segmentation.
VOC semantic evaluation uses the 1,449 validation images and 21 classes, including background.
\TAGI{} uses the 1,464-image VOC training split; the quoted semantic methods use the augmented 10,582-image set.
Instance results use the 20 foreground categories.
We report both mask AP averaged over IoU thresholds $0.50{:}0.05{:}0.95$ and AP$_{50}$, since VOC weakly supervised instance methods commonly report the latter.

\paragraph{COCO evaluation protocols.}
The semantic methods in \textcolor{mylinkcolor}{Table~\ref{tab:weak-semantic-voc-coco}} use \emph{COCO-Object}: 80 thing classes plus a background class.
Published semantic results use COCO 2014; our evaluation uses the 118,287/5,000-image COCO 2017 train/validation split, with the two versions reported in separate columns.
The panoptic comparison with TRACE and point-supervised methods uses \emph{COCO panoptic 2017}, with 80 thing and 53 stuff classes.
For \TAGI{}, COCO-Object supplies image-level training labels for object-semantic and instance evaluation, and the 171-category COCO-Stuff vocabulary supplies image-level training labels for the COCO-133 panoptic evaluation.

\paragraph{Supplementary training settings.}
VOC, COCO-Object, and COCO-Stuff use the same training configuration as the main benchmarks in \textcolor{mylinkcolor}{Table~\ref{tab:tag-training}}.
The peak learning rate is $5{\times}10^{-3}$, the training budget is 10 epochs, and $(\lambda_{\mathrm{pres}},\lambda_{\mathrm{rank}},m,r,G)=(4,0.5,3,16,16)$, without dataset-specific tuning.

\subsubsection{Segmentation Gains on VOC and COCO}
\textcolor{mylinkcolor}{Table~\ref{tab:tag-voc-coco}} compares \TAGI{} with frozen SAM3 on the complete VOC and COCO-Object validation sets.
Both instance evaluations retain at most 100 detections per image and use predicted category-presence scores, without GT-present class lists.

\begin{table}[!htbp]
\apptablestyle{7pt}
\caption{\textbf{SAM3 and \TAGI{} results on VOC and COCO-Object.}
AP denotes mask average precision over IoU thresholds $0.50{:}0.05{:}0.95$; AP$_{50}$ uses IoU $0.50$.
VOC PQ uses 20 thing classes plus background.
COCO PQ uses \emph{81 classes}, merging all stuff into background; \textcolor{mylinkcolor}{Table~\ref{tab:weak-voc-coco}} reports the 133-class evaluation.
Bold marks the higher score within each protocol.}
\label{tab:tag-voc-coco}
\begin{tabularx}{0.92\linewidth}{l!{\vrule width 0.6pt}*{4}{>{\centering\arraybackslash}X}}
\toprule
\rowcolor{apptablehead}
\apphead{Method} & \apphead{mIoU} & \apphead{PQ} & \apphead{AP} & \apphead{AP$_{50}$} \\
\midrule
\rowcolor{apptablegroup}
\multicolumn{5}{l}{\appgroup{VOC 2012 (1,449 images)}} \\
\rowcolor{white}
Frozen SAM3 & 83.49 & 81.38 & 57.34 & 80.92 \\
\rowcolor{oursrow}
\tagI{} & \textbf{89.46} & \textbf{85.94} & \textbf{59.58} & \textbf{83.61} \\
\midrule
\rowcolor{apptablegroup}
\multicolumn{5}{l}{\appgroup{COCO-Object 2017 (5,000 images)}} \\
\rowcolor{white}
Frozen SAM3 & 63.49 & 57.62 & 43.85 & 68.26 \\
\rowcolor{oursrow}
\tagI{} & \textbf{67.58} & \textbf{61.22} & \textbf{46.56} & \textbf{71.03} \\
\bottomrule
\end{tabularx}
\end{table}

The COCO PQ evaluation in \textcolor{mylinkcolor}{Table~\ref{tab:tag-voc-coco}} merges the 53 stuff categories into a single background class while preserving the 80 thing classes and ignored pixels.
The 57.62 PQ for frozen SAM3 and 61.22 PQ for \TAGI{} are therefore \emph{81-class diagnostic results}.
We report the standard 133-class evaluation separately in \textcolor{mylinkcolor}{Table~\ref{tab:weak-voc-coco}}.
\TAGI{} improves every metric in \textcolor{mylinkcolor}{Table~\ref{tab:tag-voc-coco}}.
On VOC, it gains 5.97 mIoU, 4.56 PQ, 2.24 AP, and 2.69 AP$_{50}$ points over frozen SAM3; the corresponding COCO-Object gains are 4.09, 3.60, 2.71, and 2.77 points.
These consistent improvements demonstrate that image-level adaptation strengthens semantic, panoptic, and instance prediction on both benchmarks.

\subsubsection{Comparison with Weakly Supervised Methods}
\textcolor{mylinkcolor}{Table~\ref{tab:weak-semantic-voc-coco}} adds WeCLIP~\citep{zhang2024frozen}, S2C~\citep{kweon2024sam}, and the CLIP/SAM-assisted framework of \citet{yang2024foundation}.
S2C transfers SAM knowledge into class activation maps before training a semantic segmenter; the latter framework learns CLIP prompts, uses SAM-assisted seeds, and trains a separate segmentation network.
\textcolor{mylinkcolor}{Table~\ref{tab:weak-voc-coco}} additionally compares image-level panoptic methods JTSM and DHR+TRACE, point-supervised PSPS/Point2Mask/EPLD, and instance methods BESTIE, CIM, WeakSAM, and WISH.
For WISH, we quote the image-tag-only setting.
BESTIE and CIM use their Mask R-CNN refinement results.
TRACE's quoted panoptic results combine its diffusion-derived instance edges with DHR semantics.

\begin{table}[!htbp]
\apptablestyle{4pt}
\caption{\textbf{Image-level semantic segmentation comparisons (mIoU).}
Prior denotes the foundation model used by the pipeline; backbone denotes the segmentation backbone.
All adapted methods use image-level target labels.
The COCO-Object columns contain 80 foreground classes plus background; COCO 2014 and 2017 are reported separately by evaluation split.
VOC training-set sizes are specified in the text. Bold marks the highest score in each column.}
\label{tab:weak-semantic-voc-coco}
\resizebox{\columnwidth}{!}{%
\begin{tabular}{lll!{\vrule width 0.6pt}ccc}
\toprule
\rowcolor{apptablehead}
\apphead{Method} & \apphead{Prior} & \apphead{Backbone} & \apphead{\shortstack{VOC 2012\\val}} & \apphead{\shortstack{COCO 2014\\val}} & \apphead{\shortstack{COCO 2017\\val}} \\
\midrule
\rowcolor{white}
WeCLIP~\citep{zhang2024frozen} & CLIP & ViT-B/16 & 76.4 & 47.1 & -- \\
\rowcolor{apptablestripe}
S2C~\citep{kweon2024sam} & SAM & WRN-38 & 78.2 & 49.8 & -- \\
\rowcolor{white}
Yang et al.~\citep{yang2024foundation} & CLIP + SAM & R101 & 77.3 & 48.6 & -- \\
\rowcolor{apptablestripe}
Yang et al.~\citep{yang2024foundation} & CLIP + SAM & Swin-L & 82.6 & \textbf{55.4} & -- \\
\midrule
\rowcolor{white}
Frozen SAM3 & SAM3 & SAM3 & 83.49 & -- & 63.49 \\
\rowcolor{oursrow}
\tagI{} & SAM3 & SAM3 & \textbf{89.46} & -- & \textbf{67.58} \\
\bottomrule
\end{tabular}}
\end{table}

\begin{table}[!htbp]
\apptablestyle{3.5pt}
\caption{\textbf{Weakly supervised panoptic and instance segmentation.}
PQ follows VOC-panoptic (20 things plus background) and COCO-panoptic 2017 (133 classes).
Instance columns report VOC AP$_{50}$ and COCO 2017 mask AP@[.5:.95].
Sup.\ follows \textcolor{mylinkcolor}{Table~\ref{tab:universality}}; prior denotes the foundation model used.
A dash means not reported.
Our PQ results use complete validation sets and the same inference procedure for frozen and adapted SAM3.
Published methods retain their original training and evaluation settings.
Bold marks the highest score in each metric column.}
\label{tab:weak-voc-coco}
\resizebox{\columnwidth}{!}{%
\begin{tabular}{llcl!{\vrule width 0.6pt}rr!{\vrule width 0.4pt}rr}
\toprule
\rowcolor{apptablehead}
 & & & & \multicolumn{2}{c!{\vrule width 0.4pt}}{\apphead{Panoptic PQ}} & \multicolumn{2}{c}{\apphead{Instance precision}} \\
\rowcolor{apptablehead}
\apphead{Method} & \apphead{Sup.} & \apphead{Prior} & \apphead{Backbone} & \apphead{VOC} & \apphead{COCO-133} & \apphead{VOC AP$_{50}$} & \apphead{COCO AP} \\
\midrule
\rowcolor{apptablegroup}
\multicolumn{8}{l}{\appgroup{Image-level panoptic methods}} \\
\rowcolor{white}
JTSM~\citep{shen2021toward} & \supI & -- & R18-WS & 39.0 & 5.3 & 44.2 & 6.1 \\
\rowcolor{apptablestripe}
DHR+TRACE~\citep{jo2026trace} & \supI & SD & R50 & 56.9 & 32.8 & -- & -- \\
\rowcolor{white}
DHR+TRACE~\citep{jo2026trace} & \supI & SD & Swin-L & 69.8 & 43.1 & -- & -- \\
\midrule
\rowcolor{apptablegroup}
\multicolumn{8}{l}{\appgroup{Point-supervised panoptic methods}} \\
\rowcolor{white}
PSPS~\citep{fan2022pointly} & \supPt & -- & R50 & 49.8 & 29.3 & -- & -- \\
\rowcolor{apptablestripe}
Point2Mask~\citep{li2023point2mask} & \supPt & -- & Swin-L & 61.0 & 37.0 & -- & -- \\
\rowcolor{white}
EPLD~\citep{li2024learnable} & \supPt & -- & Swin-L & 68.5 & 41.0 & -- & -- \\
\midrule
\rowcolor{apptablegroup}
\multicolumn{8}{l}{\appgroup{Image-level instance methods}} \\
\rowcolor{white}
BESTIE~\citep{kim2022beyond} & \supI & -- & HRNet-48 & -- & -- & 51.0 & 14.3 \\
\rowcolor{apptablestripe}
CIM~\citep{li2023complete} & \supI & -- & R50 & -- & -- & 55.9 & 17.0 \\
\rowcolor{white}
WeakSAM~\citep{zhu2024weaksam} & \supI & SAM & R50 + M2F & -- & -- & 64.4 & 25.2 \\
\rowcolor{apptablestripe}
WISH~\citep{kweon2025wish} & \supI & SAM & R50 & -- & -- & 62.9 & 20.7 \\
\midrule
\rowcolor{apptablegroup}
\multicolumn{8}{l}{\appgroup{SAM3 baseline and adaptation}} \\
\rowcolor{white}
Frozen SAM3 & \supN & SAM3 & SAM3 & 81.38 & 49.18 & 80.92 & 43.85 \\
\rowcolor{oursrow}
\tagI{} & \supI & SAM3 & SAM3 & \textbf{85.94} & \textbf{51.57} & \textbf{83.61} & \textbf{46.56} \\
\bottomrule
\end{tabular}}
\end{table}

\Needspace{6\baselineskip}
\paragraph{Comparison with SAM-based pipelines.}
On VOC, \TAGI{} obtains 89.46 mIoU, exceeding S2C by 11.26 points and the strongest quoted CLIP/SAM-assisted configuration by 6.86 points.
Its 83.61 AP$_{50}$ exceeds WeakSAM and image-tag-only WISH by 19.21 and 20.71 points, respectively.
On COCO 2017, its 46.56 mask AP exceeds their 25.2 and 20.7 AP by 21.36 and 25.86 points.
The higher semantic and instance scores demonstrate the effectiveness of adapting SAM3's text-conditioned recognition with image-level supervision.
Among the compared image-level pipelines, \TAGI{} is the higher-accuracy choice for VOC semantic segmentation and for instance segmentation on both VOC and COCO.

\paragraph{Panoptic performance and comparison with TRACE.}
\TAGI{} achieves the highest panoptic scores in \textcolor{mylinkcolor}{Table~\ref{tab:weak-voc-coco}}: 85.94 PQ on VOC and 51.57 PQ on COCO-133.
It exceeds the strongest DHR+TRACE configuration by 16.14 and 8.47 PQ points, respectively, under image-level supervision.
It also surpasses all listed point-supervised methods on both benchmarks, including EPLD by 17.44 and 10.57 PQ points.
With matched panoptic inference, the gains over frozen SAM3 are 4.56 and 2.39 PQ points, demonstrating the benefit of \TAG{} adaptation on the shared foundation model.
These results establish \TAGI{} as a higher-accuracy option among the evaluated weakly supervised panoptic methods and support reusing pretrained spatial priors through image-level recognition calibration.

\Needspace{4\baselineskip}
\subsection{Comparison with SAM3 Calibration Methods}\label{sec:appendix-sam3-calibration}

\paragraph{Benchmarks and counterparts.}
We compare \TAG{} with ConceptBank~\citep{conceptbank}, CoCo-SAM3~\citep{chen2026cocosam3}, ActiveSAM~\citep{tien2026activesam}, and ProC-SAM3~\citep{song2026prompt} in \textcolor{mylinkcolor}{Table~\ref{tab:sam3-calibration}}.
We report mIoU on ADE20K, Cityscapes, iSAID, and Pascal Context-59 (PC59)~\citep{voc_context}; PC59 evaluates 59 categories, excluding background.
These methods calibrate SAM3's prompts, class selection, or prediction scores for inference without gradient-based training.
Their supervision requirements nevertheless differ: \emph{ConceptBank uses ground-truth masks from the target training split} to construct its concept bank offline through mask-pooled visual prototypes, providing spatial supervision unavailable to either \TAG{} variant.

\paragraph{Result sources and evaluation protocols.}
ConceptBank's scores are from Tables~1--2 of \citet{conceptbank}; CoCo-SAM3 and ActiveSAM's natural-image scores are from Table~1 of their respective papers.
ProC-SAM3 and the marked ActiveSAM iSAID scores come from Table~I of \citet{song2026prompt}; the latter is their reproduction.
The three main-benchmark \TAG{} scores reuse \textcolor{mylinkcolor}{Table~\ref{tab:universality}}.
Each quoted method retains its published prompt construction and inference settings.
Unreported entries are marked with a dash.

\begin{table}[!htbp]
\centering
\small
\setlength{\abovecaptionskip}{0pt}
\caption{\textbf{Comparison with SAM3 calibration methods (mIoU, \%).}
PC59: Pascal Context-59. Target labels specify the supervision used by each method; ConceptBank uses training-set GT masks.
Published results retain their original inference protocols.
$^{\ast}$: ActiveSAM reproduced by \citet{song2026prompt}; --: not reported.
\textbf{Bold} and \underline{underline} mark the best and second-best scores in each column.
$\uparrow$ gives each \TAG{} variant's gain in mIoU points over the best non-\TAG{} result in that column.}
\label{tab:sam3-calibration}
\vspace{2pt}
\setlength{\tabcolsep}{4pt}
\renewcommand{\arraystretch}{1.15}
\begin{tabular}{p{\dimexpr.30\linewidth-12\tabcolsep\relax}p{.14\linewidth}*{4}{>{\centering\arraybackslash}p{.14\linewidth}}}
\noalign{\hrule height 1.2pt}
\rowcolor{gray!20}
\textbf{Method} & \textbf{Target labels} & \textbf{ADE20K} & \textbf{Cityscapes} & \textbf{iSAID} & \textbf{PC59} \\
\hline\hline
\rowcolor{white}
ConceptBank & GT masks & 43.3 & 75.1 & 35.4 & 63.0 \\
\rowcolor{gray!10}
CoCo-SAM3 & None & 38.3 & 70.7 & -- & 61.2 \\
\rowcolor{white}
ActiveSAM & None & 40.0 & 69.7 & $28.3^{\ast}$ & 60.1 \\
\rowcolor{gray!10}
ProC-SAM3 & None & -- & -- & 29.5 & -- \\
\hline
\rowcolor{oursrowsoft}
\tagV{} & MLLM labels & \snd{47.0}\,\redup{3.7} & \snd{78.1}\,\redup{3.0} & \snd{57.4}\,\redup{22.0} & \snd{65.2}\,\redup{2.2} \\
\rowcolor{oursrow}
\tagI{} & Image labels & \best{52.7}\,\redup{9.4} & \best{80.3}\,\redup{5.2} & \best{61.4}\,\redup{26.0} & \best{69.6}\,\redup{6.6} \\
\noalign{\hrule height 1.2pt}
\end{tabular}
\end{table}

\paragraph{Beyond inference-time calibration.}\label{sec:results-sam3-calibration}
\TAGI{} and \TAGV{} achieve the best and second-best mIoU, respectively, on all four benchmarks.
\TAGI{} exceeds the strongest non-\TAG{} result by 9.4, 5.2, 26.0, and 6.6 points on ADE20K, Cityscapes, iSAID, and PC59, respectively, reaching 69.6 mIoU on PC59.
More notably, \TAGV{} surpasses every available calibration counterpart even when trained on unlabeled target images using only image-level labels generated by an off-the-shelf MLLM.
Its corresponding gains are 3.7, 3.0, 22.0, and 2.2 points, despite ConceptBank's access to target masks.
On iSAID, \TAGV{} also outperforms the remote-sensing-specific ProC-SAM3 by 27.9 points (57.4 vs.\ 29.5).
The consistent lead across all four benchmarks demonstrates the accuracy advantage of \TAG{} over the evaluated SAM3 calibration methods.
For applications with unlabeled target images and a known class vocabulary, the comparison favors \TAGV{} over the evaluated inference-time calibration methods when segmentation accuracy is the priority.

\Needspace{4\baselineskip}
\begingroup
\raggedbottom
\setlength{\intextsep}{3pt}
\setlength{\abovecaptionskip}{4pt}
\subsection{Learned Conditioning versus Retrieved Class Descriptions}
\label{sec:appendix-phrase-decoding}

Concept expansion enriches the language supplied to a segmentor~\citep{conceptbank,zhu2026vip}.
We examine the benefit of learning the text-conditioned pathway by comparing \TAGI{} with nearest-phrase substitution in frozen SAM3.
The evaluation covers six classes from ADE20K, Cityscapes, and iSAID.

\paragraph{Retrieval and evaluation.}
We retrieve alternative descriptions from a fixed bank of 15,057 phrases using cosine scores between normalized, mean-pooled phrase features and corrected text anchors, averaged over 23 discovery images.
\textcolor{mylinkcolor}{Table~\ref{tab:phrase-retrieval}} lists the nearest alternatives after excluding the exact class name.
Each substitution replaces the target class's full text encoding in frozen SAM3, preserving all competing classes and output labels.
We evaluate mean target IoU on 12 positive validation images per class, disjoint from discovery; iSAID also uses disjoint source images.
For this semantic comparison, \TAGI{} uses text, fusion, and presence adaptation.

\begin{table}[!htbp]
  \centering
  \caption{\textbf{Nearest alternative phrases.} Exact names excluded; $\dagger$: original name is top-1.}
  \label{tab:phrase-retrieval}
  \footnotesize
  \setlength{\tabcolsep}{4pt}
  \setlength{\extrarowheight}{0pt}
  \setlength{\aboverulesep}{1pt}
  \setlength{\belowrulesep}{1pt}
  \renewcommand{\arraystretch}{1.0}
  \arrayrulecolor{black}
  \begin{tabular*}{\textwidth}{@{\extracolsep{\fill}}lll@{}}
    \toprule
    \textbf{Dataset} & \textbf{Class name} & \textbf{Nearest alternative phrase} \\
    \midrule
    Cityscapes & train$^{\dagger}$ & tram \\
               & rider$^{\dagger}$ & rider in the foreground \\
    ADE20K & table$^{\dagger}$ & table furniture \\
           & cushion$^{\dagger}$ & pillow \\
    iSAID & baseball diamond & baseball diamond next to other objects \\
          & ground track field & ground track field in bright daylight \\
    \bottomrule
  \end{tabular*}
\end{table}

\Needspace{4\baselineskip}
\paragraph{Consistent gains from learned conditioning.}
\TAGI{} achieves the highest mean target IoU for all six classes in \textcolor{mylinkcolor}{Figure~\ref{fig:phrase-function}a}.
Relative to the original class names, its gains range from 32.0 to 58.3 IoU points, whereas five of the six retrieved alternatives yield gains below one point.
These results demonstrate the advantage of learned text-conditioned adaptation over nearest-phrase substitution across the evaluated categories.

\begin{figure}[!htbp]
  \centering
  \includegraphics[width=\textwidth,trim=3bp 2bp 1bp 1bp,clip]{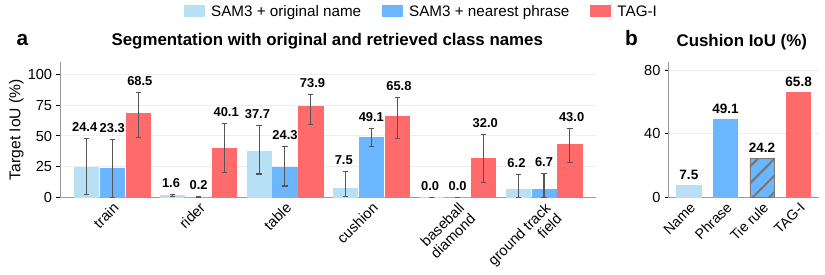}
  \caption{\textbf{Segmentation with names, retrieved phrases, and \TAGI{}.}
  (a) Mean target IoU; whiskers are 95\% intervals from 2,000 image-bootstrap resamples.
  (b) Cushion IoU when exact ties are reassigned to the existing pillow class (hatched), with model scores fixed.}
  \label{fig:phrase-function}
\end{figure}

\paragraph{Resolving closely related categories.}
Cushion is the strongest phrase-substitution result: replacing its prompt with \emph{pillow} raises IoU from 7.5 to 49.1, while \TAGI{} reaches 65.8 using the canonical class name.
Because \emph{pillow} is also an ADE20K category, the replacement creates identical competing prompts.
Assigning exact ties to the existing pillow class gives 24.2 cushion IoU with model scores fixed (\textcolor{mylinkcolor}{Figure~\ref{fig:phrase-function}b}).
\TAGI{} exceeds both substitution settings while retaining distinct category prompts.
Together, these comparisons support learning how class anchors condition segmentation to improve target-category recognition.
\endgroup

\Needspace{4\baselineskip}
\subsection{Diagnostic Experiments}\label{sec:appendix-diagnosis}

This section complements the qualitative examples and oracle experiments in \textcolor{mylinkcolor}{Section~\ref{sec:failure}}.
We connect object-level recognition errors, recoverable segmentation accuracy, and the spatial quality of recognized objects to the design of \TAG{}.

\subsubsection{Object-Level Error Structure}\label{sec:appendix-error-structure}

\paragraph{Diagnostic setup.}
We analyze frozen SAM3 predictions on ADE20K, Cityscapes, and iSAID.
An object is an 8-connected GT foreground component with at least 20 valid pixels; ignored pixels are excluded, and touching instances of the same class may form one component.
We regard an object as \emph{failed} when fewer than half of its pixels receive the correct category label.
This criterion measures correct-class pixel recall within each GT object.
For each failed object, we compute the fraction of its pixels assigned an incorrect label and the fraction assigned its most frequent predicted label, including background where applicable.
Both statistics give each failed object equal weight, irrespective of its area.
Table~\ref{tab:diagnostic-coverage} records the evaluation populations.

\begin{table}[!htbp]
  \apptablestyle{5pt}
  \caption{Evaluation populations for the object-level spatial statistics. iSAID observations are validation tiles. Failed objects have correct-class pixel recall below $0.5$.}
  \label{tab:diagnostic-coverage}
  \begin{tabularx}{\linewidth}{l*{3}{>{\centering\arraybackslash}X}}
    \toprule
    \rowcolor{apptablehead}
    \apphead{Dataset} & \apphead{Images / tiles} & \apphead{GT objects} & \apphead{Failed objects} \\
    \midrule
    ADE20K & 2{,}000 & 40{,}740 & 16{,}281 \\
    Cityscapes & 500 & 46{,}490 & 19{,}225 \\
    iSAID & 11{,}644 & 229{,}354 & 130{,}650 \\
    \bottomrule
  \end{tabularx}
\end{table}

\begin{figure}[!htbp]
  \centering
  \includegraphics[width=\textwidth,trim=5bp 13bp 6bp 6bp,clip]{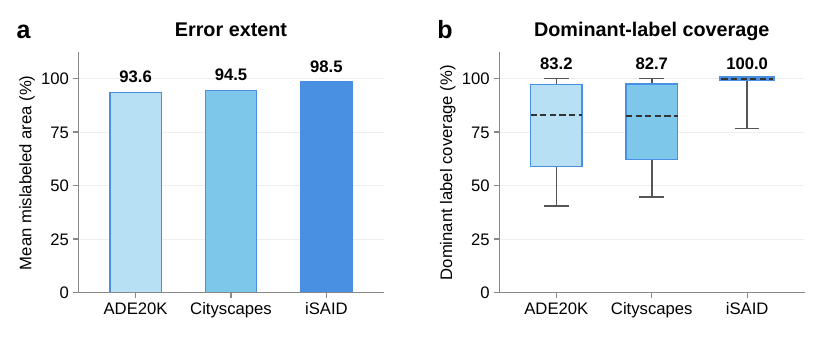}
  \caption{\textbf{Object-level structure of frozen SAM3's recognition errors.}
  Failed objects are GT foreground connected components with correct-class pixel recall below 0.5.
  \textbf{(a)} Mean fraction of mislabeled pixels within each failed object.
  \textbf{(b)} Fraction of each failed object covered by its most frequent predicted label, including background where applicable.
  Boxes and whiskers denote the interquartile range and 5th--95th percentiles; dashed lines mark medians, and the iSAID box collapses at 100\%.
  Numerical labels give means in (a) and medians in (b).
  Each failed object contributes equally to both summaries.}
  \label{fig:failure-statistics}
\end{figure}

\Needspace{4\baselineskip}
\paragraph{Errors are extensive and often dominated by one label.}
Figure~\ref{fig:failure-statistics}a reports mean mislabeled fractions of $93.6\%$, $94.5\%$, and $98.5\%$ on ADE20K, Cityscapes, and iSAID, respectively.
The high dominant-label medians in Figure~\ref{fig:failure-statistics}b indicate that many failed objects receive a consistent wrong category or background label across most of their pixels.
The errors therefore frequently extend across the object interior, highlighting category recognition and concept activation as useful targets for adaptation.
The oracle interventions below test how much segmentation accuracy can be recovered by correcting these predictions while reusing SAM3's spatial outputs.

\subsubsection{Oracle Interventions and Spatial Quality}\label{sec:appendix-oracle-recovery}

\paragraph{Oracle definitions.}
The geometry reference, \emph{GT mask regions}, supplies ground-truth regions while retaining SAM3's category scores.
The category and activation interventions in \textcolor{mylinkcolor}{Figure~\ref{fig:failure-summary}b} keep SAM3's raw masks, semantic maps, and objectness logits fixed.
\emph{GT categories} assigns each 8-connected foreground component of the final prediction its majority GT label over valid pixels, including background, while preserving its pixel support.
Ties use total GT class area and then class index; ignored-only components retain their labels.
\emph{Positive gates} sets presence scores of GT-present classes to one when scoring and filtering instances and during final fusion, including the explicit background class on iSAID; GT-absent classes retain their original scores.
\emph{Gates \& categories} applies positive activation followed by majority-GT naming of the resulting foreground components.
GT annotations construct these diagnostics; \TAG{} inference uses class names and predicted presence scores.

\paragraph{Recognition correction recovers segmentation accuracy.}
As shown in \textcolor{mylinkcolor}{Figure~\ref{fig:failure-summary}b}, assigning GT categories to predicted regions improves ADE20K by 47.2 mIoU points over frozen SAM3.
Positive activation improves all three benchmarks, and combining category and activation corrections further increases accuracy.
These gains demonstrate that better concept recognition can turn existing spatial predictions into substantially more accurate segmentation, directly supporting the motivation for \TAG{}.

\Needspace{15\baselineskip}
\begin{wrapfigure}{r}{0.53\textwidth}
  \vspace{-5pt}
  \centering
  \includegraphics[width=\linewidth]{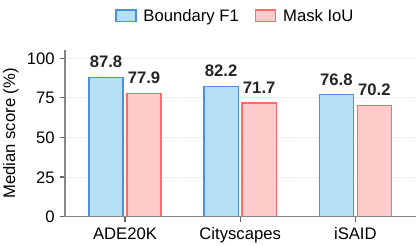}
  \par\vspace{-5pt}
  \setlength{\abovecaptionskip}{2pt}
  \fontsize{8}{9.5}\selectfont
  \caption{\textbf{Spatial quality of recognized objects.}
  Median boundary F1 at $5$\,px tolerance and mask IoU (\%).}
  \label{fig:recognized-geometry}
  \vspace{-3pt}
\end{wrapfigure}

\paragraph{Spatial quality of recognized objects.}
Recognized objects have correct-class recall at least $0.5$; both spatial metrics use same-class predictions within the GT component bounding box expanded by $10$\,px, excluding ignored pixels.
Figure~\ref{fig:recognized-geometry} reports median boundary F1 at a $5$\,px tolerance~\citep{davis,cheng2021boundary} and median mask IoU for this population.
Median mask IoU is $77.9\%$, $71.7\%$, and $70.2\%$ on ADE20K, Cityscapes, and iSAID, respectively; median boundary F1 ranges from $76.8\%$ to $87.8\%$.
These measurements show useful boundary and region information for recognized objects across the three domains.

The diagnostics identify recoverable recognition errors alongside useful pretrained spatial information.
\TAG{} builds on these findings by using image-level labels to learn text-conditioned corrections and calibrate presence, while reusing SAM3's pretrained mask heads for dense prediction.
The segmentation gains in \textcolor{mylinkcolor}{Table~\ref{tab:universality}} demonstrate the effectiveness of this adaptation strategy across all three domains and tasks.
\par\WFclear

\Needspace{4\baselineskip}
\subsection{New-Vocabulary Evaluation on ADE847}\label{sec:appendix-new-vocabulary}

\paragraph{Evaluation protocol.}
To complement the illustrative categories in \textcolor{mylinkcolor}{Figure~\ref{fig:general}b}, we evaluate ADE150-to-ADE847 transfer on all 2,000 validation images with all 847 labels competing at inference.
We transfer the ADE150-trained \TAGI{} adapters directly, re-encode the target class names, and use a single-view semantic readout with predicted presence scores.
All transferred parameters remain fixed during this evaluation.
Ground-truth class lists are never supplied at inference.
We accumulate pixel intersections and unions over the validation set before averaging class IoUs.
The 150 labels mapped to the training vocabulary and the remaining 697 additional labels are reported separately, while preserving competition among all 847 classes in both groups.

\paragraph{Transferred components.}
\textcolor{mylinkcolor}{Table~\ref{tab:ade847-new-vocabulary}} compares frozen SAM3, text correction alone, downstream correction alone, and their combination.
Text correction applies the trained feature adapter to the newly encoded text; downstream correction reuses the fusion, presence-token, and decoder image-K/V adapters.
The downstream-only condition uses the original text features; all conditions share the same vocabulary and semantic readout.
The four conditions select which learned adapters are active at inference.

\begin{table}[!htbp]
  \apptablestyle{5pt}
  \caption{\textbf{ADE150-to-ADE847 vocabulary expansion (mIoU, \%).}
  All conditions evaluate the complete validation set with 847 candidate labels.
  Columns partition the labels by correspondence to the ADE150 training vocabulary.
  \textbf{Bold} marks the highest score in each column.
  The last row gives the combined correction's change from frozen SAM3 in mIoU points.}
  \label{tab:ade847-new-vocabulary}
  \begin{tabularx}{\linewidth}{X*{2}{>{\centering\arraybackslash}p{.21\linewidth}}}
    \toprule
    \rowcolor{apptablehead}
    \apphead{Transferred components} & \apphead{Mapped 150} & \apphead{Additional 697} \\
    \midrule
    None (frozen SAM3) & 30.52 & 11.39 \\
    \rowcolor{apptablestripe}
    Text correction only & 45.18 & 12.68 \\
    Downstream correction only & 30.88 & 12.28 \\
    \rowcolor{apptablegroup}
    Text + downstream correction & \textbf{45.45} & \textbf{13.32} \\
    \midrule
    Change from frozen SAM3 & $+14.93$ & $+1.93$ \\
    \bottomrule
  \end{tabularx}
\end{table}

\paragraph{Generalization to additional labels.}
Combining text and downstream correction improves mIoU on the 697 additional labels from 11.39 to 13.32, a gain of 1.93 points over frozen SAM3.
Text correction alone reaches 12.68 mIoU, while downstream correction alone reaches 12.28, corresponding to gains of 1.29 and 0.89 points.
Their combination exceeds either component alone, indicating that the two correction pathways contribute complementary benefits to vocabulary expansion.
The 150 mapped labels also improve from 30.52 to 45.45 mIoU, a gain of 14.93 points under full-vocabulary competition.
The improvements in both groups show that \TAG{} benefits the expanded ADE vocabulary while retaining strong recognition of the mapped training labels.
Together with \textcolor{mylinkcolor}{Figure~\ref{fig:general}b}, the average gain over 697 additional labels demonstrates the utility of reusing learned corrections beyond the training vocabulary.

\clearpage
\section{Additional Qualitative Results}\label{sec:appendix-qualitative}

\textcolor{mylinkcolor}{Figures~\ref{fig:qualitative-city}--\ref{fig:qualitative-isaid}} compare panoptic predictions on eight additional validation scenes per benchmark, illustrating improved category consistency and object coverage after \TAG{} adaptation.

\begin{figure}[!ht]
  \centering
  \includegraphics[width=\textwidth,trim=0bp 5bp 0bp 0bp,clip]{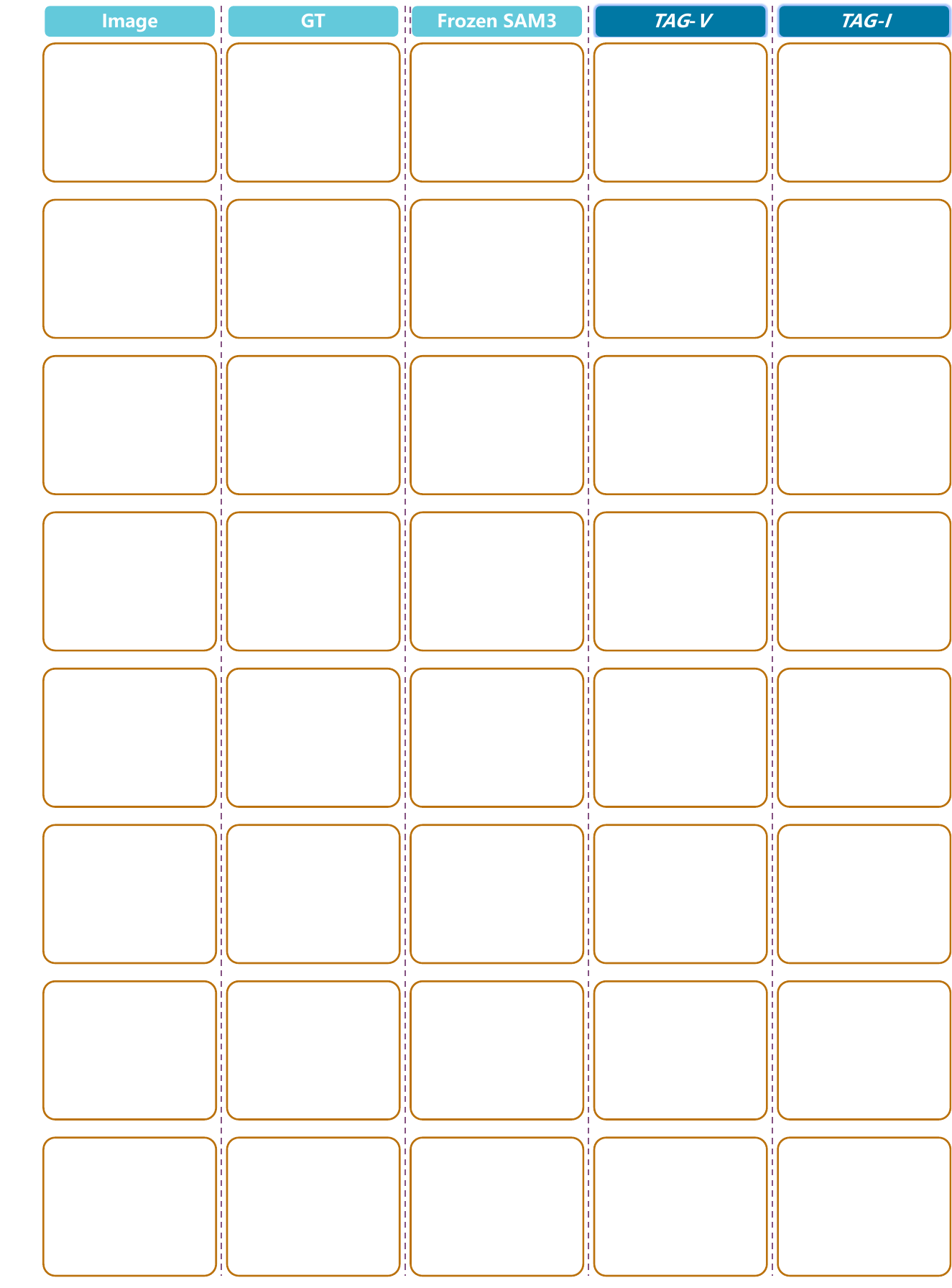}
  \caption{\textbf{Additional panoptic results on Cityscapes.} Columns show the image, ground truth, frozen SAM3, \TAGV{}, and \TAGI{}. \TAG{} improves person/rider and bus/train discrimination and recovers vegetation regions missed by frozen SAM3. All columns use the same crop and category palette.}
  \label{fig:qualitative-city}
\end{figure}

\clearpage
\begin{figure}[!ht]
  \centering
  \includegraphics[width=\textwidth,trim=0bp 5bp 0bp 0bp,clip]{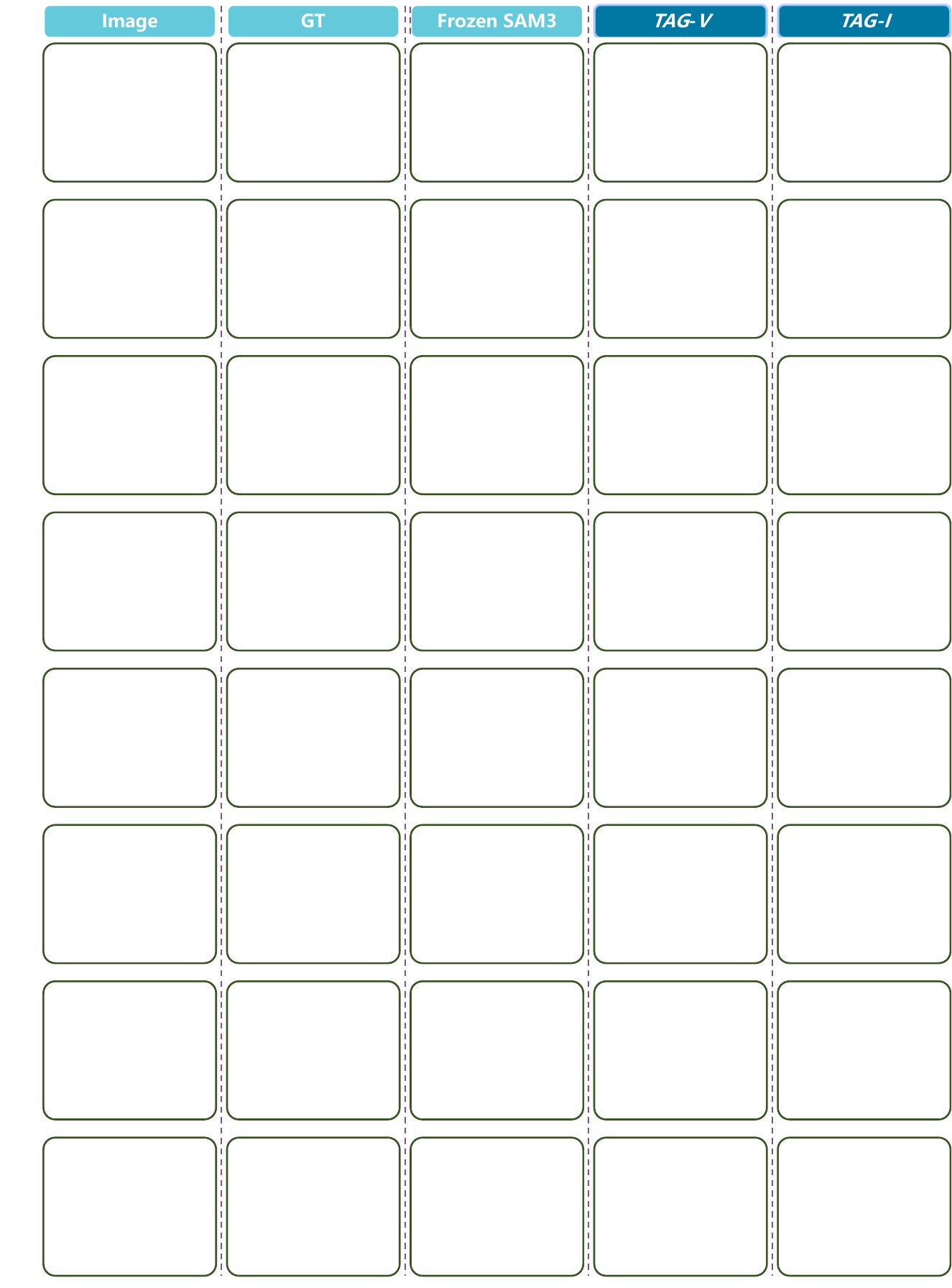}
  \caption{\textbf{Additional panoptic results on ADE20K.} \TAG{} improves recognition of cushions and furniture categories and recovers regions missed by frozen SAM3, including cradles and kitchen islands. The indoor and outdoor scenes illustrate gains in category consistency and region coverage.}
  \label{fig:qualitative-ade}
\end{figure}

\clearpage
\begin{figure}[!ht]
  \centering
  \includegraphics[width=\textwidth,trim=0bp 5bp 0bp 0bp,clip]{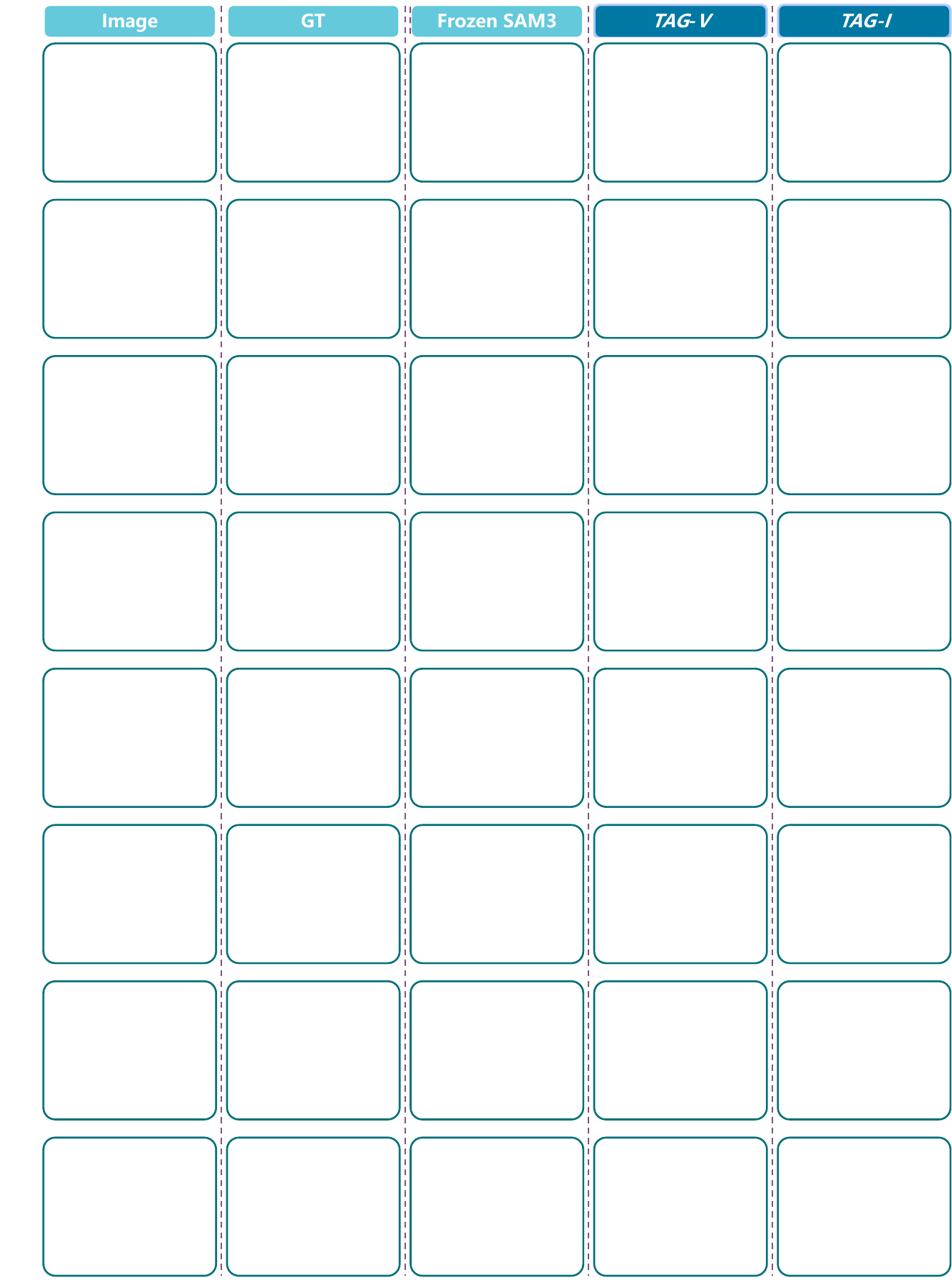}
  \caption{\textbf{Additional panoptic results on iSAID.} \TAG{} improves court/field discrimination, recovers storage tanks and athletics facilities, and reduces foreground leakage around piers. The last two rows show predictions for crowded boats and small vehicles.}
  \label{fig:qualitative-isaid}
\end{figure}

\clearpage
\section{Dataset-Specific Prompts for Presence Labeling}\label{sec:appendix-prompts}

The following condensed templates retain representative category rules.
Each example entry is \texttt{[answer, probability]}: a yes/no decision followed by the token-derived probability $p_k\in[0,1]$ of category presence (\textcolor{mylinkcolor}{Sec.~\ref{sec:vlm}}).
CWD converts $p_k$ into training targets and confidence weights using \textcolor{mylinkcolor}{Eqs.~\ref{eq:smoothing} and~\ref{eq:cwd}}.
A short description summarizes the input image for each example.

\definecolor{presencepromptblue}{HTML}{247DB3}
\definecolor{presencepromptpale}{HTML}{F3F8FC}
\tcbset{presence prompt/.style={
  enhanced,
  colback=presencepromptpale, colframe=presencepromptblue,
  colbacktitle=presencepromptblue, coltitle=white,
  fonttitle=\small\sffamily, center title,
  boxrule=0.65pt, arc=1.2mm,
  left=8pt, right=8pt, top=4pt, bottom=4pt,
  before skip=4pt, after skip=4pt,
  listing only, listing engine=listings,
  listing options={
    basicstyle=\fontencoding{T1}\fontfamily{cmtt}\fontsize{9}{11.5}\selectfont,
    columns=fullflexible, keepspaces=true,
    breaklines=true, breakatwhitespace=true,
    breakindent=1em, showstringspaces=false,
    aboveskip=0pt, belowskip=0pt
  }
}}

\subsection*{ADE20K: Indoor and Outdoor Scenes}
The field \texttt{\{category\_list\}} denotes the supplied ADE20K categories; the full vocabulary has 150 classes.
Only representative names, answers, and scores are displayed.

\begin{tcblisting}{presence prompt, title={ADE20K: condensed presence-labeling prompt}}
This is a photograph of an indoor or outdoor scene. Identify the categories visible in the image.

Candidate categories: {category_list}
Examples: wall, floor, ceiling, bed, windowpane, curtain, pillow, lamp, chair, sofa, person, ...

Rules:
1. Inspect the entire image, including both objects and visible background surfaces.
2. Judge presence from visual evidence. Scene context may help, but a typical object need not be present.
3. Include small, distant, partly occluded, or truncated objects when they remain recognizable.
4. Distinguish related categories by their visual role; for example, a pillow on a bed differs from a cushion on a sofa.
5. Keep the supplied category names unchanged, and consider only categories in the candidate list.

Response format:
Return only a JSON object with one entry per supplied category and "yes" or "no" as its value.
\end{tcblisting}

\paragraph{Example input.}
An RGB bedroom image showing walls, a floor, a ceiling, a bed with pillows, a curtained window, and a bedside lamp.
The dictionary below shows selected categories; all supplied categories receive an answer and a score in the full output.

\begin{tcblisting}{presence prompt, title={ADE20K: example answers and probabilities (selected classes)}}
{
  "wall": ["yes", 0.99], "floor": ["yes", 0.98],
  "ceiling": ["yes", 0.96], "bed": ["yes", 0.99],
  "windowpane": ["yes", 0.91], "curtain": ["yes", 0.88],
  "pillow": ["yes", 0.92], "lamp": ["yes", 0.93],
  "cushion": ["no", 0.37], "chair": ["no", 0.16],
  "sofa": ["no", 0.08], "person": ["no", 0.02]
}
\end{tcblisting}

\clearpage
\subsection*{Cityscapes: Urban Street Scenes}
The field \texttt{\{category\_list\}} denotes the 19 Cityscapes categories.
The template highlights a few common distinctions, while the example includes answers and probabilities for all 19 keys.
Canonical names such as \texttt{trafficlight} and \texttt{trafficsign} are preserved.

\begin{tcblisting}{presence prompt, title={Cityscapes: condensed presence-labeling prompt}}
This is a street-level photograph taken from a vehicle. Identify the categories visible in the image.

Candidate categories: {category_list}
Examples: road, sidewalk, building, vegetation, terrain, person, rider, car, bicycle, ...

Rules:
1. Scan the whole scene, including distant objects, image edges, and partly occluded instances.
2. Distinguish pedestrians from riders. A visible cyclist and bicycle support both "rider" and "bicycle".
3. Separate low grass or exposed ground ("terrain") from trees and bushes ("vegetation").
4. Identify vehicles from visible appearance; for example, reserve "bus" for recognizable buses rather than uncertain distant vehicles.
5. Ignore the recording vehicle's hood. Use only the supplied categories and preserve their exact names.

Response format:
Return only a JSON object with one entry per supplied category and "yes" or "no" as its value.
\end{tcblisting}

\paragraph{Example input.}
An RGB street image with roads, sidewalks, buildings, trees, a narrow grass strip, traffic signals, cars, a pedestrian, and a cyclist.
The example illustrates scores for one view, including uncertainty about a small terrain region.

\begin{tcblisting}{presence prompt, title={Cityscapes: example answers and probabilities (19 classes)}}
{
  "road": ["yes", 0.99], "sidewalk": ["yes", 0.97],
  "building": ["yes", 0.98], "wall": ["no", 0.12],
  "fence": ["no", 0.24], "pole": ["yes", 0.94],
  "trafficlight": ["yes", 0.91], "trafficsign": ["yes", 0.89],
  "vegetation": ["yes", 0.95], "terrain": ["yes", 0.58],
  "sky": ["yes", 0.99], "person": ["yes", 0.93],
  "rider": ["yes", 0.87], "car": ["yes", 0.98],
  "truck": ["no", 0.18], "bus": ["no", 0.07],
  "train": ["no", 0.01], "motorcycle": ["no", 0.11],
  "bicycle": ["yes", 0.92]
}
\end{tcblisting}

\noindent Each entry pairs the yes/no answer with its cached category-presence probability.
For multiple labeling views, the per-class maximum supplies the image-level score.

\clearpage
\subsection*{iSAID: Overhead Aerial Imagery}
The field \texttt{\{category\_list\}} denotes iSAID's 15 foreground classes and \texttt{background}.
The template summarizes visual distinctions relevant to aerial imagery; the example includes all 16 keys with their canonical spellings.

\begin{tcblisting}{presence prompt, title={iSAID: condensed presence-labeling prompt}}
This is an overhead aerial image tile. Identify the categories visible from the top-down view.

Candidate categories: {category_list}
Examples: ship, harbor, store tank, bridge, large vehicle, small vehicle, plane, background, ...

Rules:
1. Inspect the full tile for small objects. Recognizable objects cut by the tile boundary still count.
2. Require visible object evidence; for example, a runway alone does not indicate "plane".
3. Distinguish related structures: ships are vessels, while harbors are dock or pier structures. Both may be present.
4. Use size and shape to separate similar categories, such as small versus large vehicles or tanks versus roundabouts.
5. Use only the supplied names. Treat roads, buildings, vegetation, and other non-target regions as background.

Response format:
Return only a JSON object with one entry per supplied category and "yes" or "no" as its value.
\end{tcblisting}

\paragraph{Example input.}
An RGB overhead port tile containing ships, piers, circular storage tanks, cars, and elongated trucks.
Roads, buildings, and open water form the background.
Small vehicles and structures can receive intermediate scores when their appearance is ambiguous.

\begin{tcblisting}{presence prompt, title={iSAID: example answers and probabilities (16 classes)}}
{
  "background": ["yes", 1.00], "ship": ["yes", 0.96],
  "store tank": ["yes", 0.88], "baseball diamond": ["no", 0.03],
  "tennis court": ["no", 0.04], "basketball court": ["no", 0.06],
  "ground track field": ["no", 0.02], "bridge": ["no", 0.23],
  "large vehicle": ["yes", 0.68], "small vehicle": ["yes", 0.84],
  "helicopter": ["no", 0.02], "swimming pool": ["no", 0.03],
  "roundabout": ["no", 0.09], "soccer ball field": ["no", 0.02],
  "plane": ["no", 0.04], "harbor": ["yes", 0.98]
}
\end{tcblisting}

\noindent Each entry pairs the yes/no answer with its cached category-presence probability.
The background score is fixed to one and excluded from presence supervision (\textcolor{mylinkcolor}{Appendix~\ref{sec:appendix-training}}).

\endgroup

\end{document}